\documentclass[acmsmall]{acmart}

\usepackage{epsfig}
\usepackage{mathtools}
\usepackage{algorithm}
\usepackage{algorithmic}
\usepackage{comment}
\usepackage{enumitem}

\AtBeginDocument{%
  \providecommand\BibTeX{{%
    \normalfont B\kern-0.5em{\scshape i\kern-0.25em b}\kern-0.8em\TeX}}}

\setcopyright{acmcopyright}
\copyrightyear{2024}
\acmYear{2024}
\acmDOI{XXXXXXX.XXXXXXX}

\begin{document}

\title{Iterated Local Search with Linkage Learning}

\author{Renato Tin\'os}
\orcid{0000-0003-4027-8851}
\affiliation{
  \institution{University of S\~ao Paulo}
  \streetaddress{Av. Bandeirantes, 3900}
  \city{Ribeir\~ao Preto} 
  \country{Brazil}
  \postcode{14040901}
}
\email{rtinos@ffclrp.usp.br}
 	
\author{Michal W. Przewozniczek}
\orcid{0000-0003-2446-6473}
\affiliation{
    \institution{Wroclaw University of Science and Technology}
    \city{Wroclaw}
    \country{Poland} 
}
\email{michal.przewozniczek@pwr.edu.pl}
	
\author{Darrell Whitley}
\orcid{0000-0002-2752-6534}
\affiliation{
  \institution{Colorado State University}
  \streetaddress{1100 Center Avenue Mall}
  \city{Fort Collins} 
  \country{USA}
  \postcode{80523-1873}
  }
\email{whitley@cs.colostate.edu}
\author{Francisco Chicano}
\orcid{0000-0003-1259-2990}
\affiliation{
  \institution{University of M\'alaga}
  \streetaddress{Bulevar Louis Pasteur, 35}
  \city{Malaga} 
  \country{Spain} 
  \postcode{29071}
}
\email{chicano@uma.es}

\renewcommand{\shortauthors}{R. Tin\'os et al.}

\begin{abstract}
In pseudo-Boolean optimization, a variable interaction graph represents variables as vertices, and interactions between pairs of variables as edges. In black-box optimization, the variable interaction graph may be at least partially discovered by using empirical linkage learning techniques. These methods never report false variable interactions, but they are computationally expensive. The recently proposed local search with linkage learning discovers the partial variable interaction graph as a side-effect of iterated local search. 
 However, information about the strength of the interactions is not learned by the algorithm. 
We propose local search with linkage learning 2, which builds a weighted variable interaction graph that stores information about the strength of the interaction between variables. 
The  weighted variable interaction graph can provide new insights about the optimization problem and behavior of optimizers.  
Experiments with NK landscapes, knapsack problem, and feature selection show that local search with linkage learning 2 is able to efficiently build weighted variable interaction graphs. 
In particular, experiments with feature selection show that the weighted variable interaction graphs can be used for visualizing the feature interactions in machine learning. 
Additionally, new transformation operators that exploit the interactions between variables can be designed. 
We illustrate this ability by proposing a new perturbation operator for iterated local search.  
\end{abstract}



\maketitle

\section{Introduction}
\label{sec:int}
\emph{Local search} is a simple strategy which is effective in a wide variety of combinatorial optimization problems~\cite{papadimitriou1998}. 
Local search systematically explores a neighborhood of the current candidate solution.  An initial solution is improved in a step-by-step manner until a local optimum is found.
The basin of attraction of a local optimum $\mathbf{x}$  is defined by the subset of candidate solutions that converge to $\mathbf{x}$ under the specific local search strategy. 

One of the most successful metaheuristics that explicitly uses local search as the main strategy is \emph{iterated local search} (ILS)~\cite{lourenco2019}. 
In ILS, local search is applied to the current solution and, after reaching a local optima (or some other criteria of solution quality is reached), perturbation is applied. 
This cycle of perturbation is repeated until a stopping criterion is met. 
The perturbations represent a form of "soft restarts" of the local search,
and maybe combined with "hard restarts" which start local search again
from a randomly selected solution.
Note that iterated local search will also generate 
a sample set of local optima.
By using an appropriate distance metric that captures some notion of nearness between local optima under local search, a neighborhood structure emerges between the local optima~\cite{ochoa2008}. 

Here, we propose a new local search strategy for ILS that iteratively discovers the structure of the problem's instance. 
Given a pseudo-Boolean optimization problem,
the strategy learns how the variables interact in the evaluation function $f(.)$.  
This strategy will be more effective when the optimization problem has bounded nonlinearity.
The structure is discovered even in problems where 
the evaluation function is hidden, for example in black-box optimization. 
Knowing the structure of the problem is useful for two main reasons. 
First, we can design specialized transformation operators and search strategies that take advantage of the knowledge of the interaction between variables to guide the search~\cite{whitley2016}. 
For example, \emph{partition crossover} uses information about the interaction between variables to design recombination masks that 
are able to decompose $f(.)$ in $k$-bounded pseudo-Boolean optimization problems~\cite{tinos2015}.  
Second, the structure can provide useful insights for understanding the nature of the problems, the behavior of algorithms, and the efficiency of operators and strategies. 
For example, some heuristics are effective when problems are separable~\cite{li2013}, i.e., when variables do not interact in $f(.)$, but are ineffective otherwise.  Knowing the variable interactions can also reveal
if the variable interactions have a bounded tree-width or if a problem appears to be submodular.

Metaheuristics should use information from the problem structure whenever possible~\cite{whitley2019}. 
In many applications, information about the structure of the problem's instance is available by inspecting the evaluation function.  
In this case, we say that the problem is a \emph{gray-box optimization problem}, in opposition to a \emph{black-box optimization problem}, i.e., a problem where this information is not available \emph{a priori}. 
However, information about the interaction between variables can be discovered in black-box optimization problems by using \emph{linkage learning} techniques. 
In fact, in some problems, information about the interaction between variables can be learned in polynomial time~\cite{heckendorn2004}. 

Linkage learning is a collection of techniques that allow the discovery of variable dependencies. Linkages may be discovered in the pre-optimization phase or during the optimization process \cite{thierens2012}. In the optimization of combinatorial problems, it is frequent to use the latter strategy \cite{goldman2014,przewozniczek2021,bosman2016}. 
Some state-of-the-art optimizers dedicated to solving discrete-encoded combinatorial problems use \emph{statistical linkage learning} \cite{goldman2014,thierens2013,hsu2015}. In statistical linkage learning, we count the frequencies of gene value combinations. On this base, we can compute the entropy and build the \emph{dependency structure matrix} that represents the predicted inter-variable dependency strengths (based on the frequency statistics). 

Statistical linkage learning can report a \textit{false linkage}, i.e., it may report a 
variable interaction that does not exist.
The curse of false linkage may significantly deteriorate the effectiveness of statistical linkage learning based optimizers \cite{przewozniczek2020,przewozniczek2020b}. 
To prevent this problem, \emph{empirical linkage learning} techniques were proposed. 
Empirical linkage learning techniques~\cite{przewozniczek2020,przewozniczek2021} are based on analyzing the differences between neighboring solutions which are generated during local search with and without perturbation. 
Proofs show that empirical linkage learning techniques never report false linkage, 
but the first empirical linkage learning methods had a very high computational cost.  
New empirical linkage learning techniques overcome these initial disadvantages and bring new, promising features. \emph{Direct linkage empirical discovery} (DLED) \cite{przewozniczek2021} discovers only direct dependencies between variables, and although it remains computationally expensive, it significantly reduces the decomposition costs.
Moreover, the hybridization of empirical linkage learning and gray-box optimization operators allows for detecting the \textit{missing linkage}, i.e., to detect that in the two groups of variables, there must be at least one pair of variables from both of these groups that are directly dependent on each other, 
and this dependency was not discovered yet. 
The missing linkage detection has led to a significant reduction of DLED-based decomposition costs because the dependency discovery is performed only when necessary (i.e., when the missing linkage is detected) \cite{przewozniczek2022}. 

\citet{tinos2022} proposed a new local search strategy for ILS that incorporates empirical linkage learning. 
In \emph{local search with linkage learning} (LSwLL), the information about the interaction between variables is stored in an empirical \emph{variable interaction graph} (VIG)~\cite{chicano2014}. 
The VIG is an undirected graph where the vertices represent decision variables and edges represent nonlinear interaction between these variables. 
The strategy used in LSwLL for building the empirical VIG is based on DLED. 
However, unlike in DLED, no additional evaluations are necessary for finding the interactions between variables. 

The VIG can be used in gray-box optimization problems to efficiently recombine solutions~\cite{tinos2015,chicano2021,tinos2021} and find the best improving moves~\cite{chicano2014}. 
However, information about the strength of the interactions is not stored in the VIG. 
Such information may be useful in solving problem instances 
when the VIG is dense~\cite{tinos2022}. 
Therefore, we propose \emph{local search with linkage learning 2} (LSwLL2), that builds an empirical \emph{weighted variable interaction graph} (VIGw).
The VIGw is a weighted undirected graph where the weights represent the strength of the nonlinear interaction between variables. 

The empirical VIGw can provide new insights about the nature of the optimization problem and behavior of the optimizers. 
Some examples are presented in this paper. 
In particular, we present experiments for the feature selection problem where the VIGw reveals how features interact in machine learning datasets. 
Additionally, new transformation operators and strategies can be designed by using the information stored in the VIGw. 
In ILS, these operators and strategies may result in speeding-up local search. 
For example, if we discover that a given decision variable does not interact with none of the other variables, we can independently optimize this variable and then keep it fixed, what should result in a speed-up for local search. 

We illustrate how information about the strength of the interaction between variables can improve ILS performance by proposing a new perturbation strategy based on the VIGw. 
Perturbation is a key element of ILS because it allows jumping between basins of attraction~\cite{lourenco2019}. 
It is also used in \emph{evolutionary algorithms} (EAs) and other metaheuristics to escape from local optima~\cite{coffin2006}. 
\citet{tinos2022} proposed a perturbation strategy based on the VIG. 
However, this strategy is not efficient when the VIG is dense. 
The perturbation strategy proposed here takes advantage of the strength of the interaction between variables to minimize this problem. 
Another advantage of the proposed strategy is that it is parameterless.

The rest of the paper is organized as follows. 
In Section~\ref{sec:background}, some definitions are presented. Then, LSwLL2 is presented in Section~\ref{sec:LSwLL}. 
The perturbation strategy based on the VIGw is presented in Section~\ref{sec:VIGwbP}. 
Experimental results are presented and analyzed in Section~\ref{sec:exp}, and the paper is concluded in Section~\ref{sec:con}.

\section{BACKGROUND}
\label{sec:background}
We are interested in pseudo-Boolean optimization, where the evaluation (fitness) function is $f: \mathbb{B}^N \rightarrow \mathbb{R}$, and $\mathbb{B}=\{0,1\}$. 
Let $f(\mathbf{x}\oplus\mathbf{1}_g)$ denote the fitness of a solution $\mathbf{x} \in \mathbb{B}^N$ after flipping the $g$-th bit,
where $\oplus$ is the XOR bitwise operation and $\mathbf{1}_g \in \mathbb{B}^N$ denotes a characteristic vector with the $g$-th element equal to one and all the other elements equal to zero. 
Using the same notation,  $f\big(\mathbf{x}\oplus(\mathbf{1}_h+\mathbf{1}_g)\big)$ is the fitness of a solution $\mathbf{x} \in \mathbb{B}^N$ after
$x_h$ and $x_g$ are flipped.

We will also denote the difference in the evaluation $f(\mathbf{x})$ when the $g$-th bit is flipped by $\delta_g(\mathbf{x})$. Thus:
\begin{equation} 
	\label{eq:deltaf3}
		\delta_g (\mathbf{x}) = f(\mathbf{x}\oplus\mathbf{1}_g)-f(\mathbf{x}). 
\end{equation}
When variable $x_g$ is flipped after flipping $x_h$, the difference is denoted by:
\begin{equation} 
	\label{eq:deltaf4}
	    \delta_{g}(\mathbf{x}\oplus\mathbf{1}_h) = f\big(\mathbf{x}\oplus(\mathbf{1}_h+\mathbf{1}_g)\big)-f(\mathbf{x}\oplus\mathbf{1}_h).
\end{equation}

Let us consider that a first-improvement local search strategy $LS(.)$ is applied for maximizing pseudo-Boolean functions according to the following definition. 

\begin{definition}
\label{def:LS}
In the local search strategy $LS(.)$: 
\begin{enumerate}[label=\roman*.]
    \item  One bit (decision variable) is flipped each time; 
    \item A tabu mechanism ensure that the $g$-th bit is flipped again only after all other bits are flipped; 
    \item Given a solution $\mathbf{x}$,  the flip in the $g$-th bit is accepted only if  $\delta_g(\mathbf{x})>0$; 
    \item  The search stops when no improvement is found.
\end{enumerate}
\end{definition}

Alg.~\ref{alg:LS} shows the pseudo-code of $LS(.)$\footnote{In the pseudo-codes,
function rand($S$) returns a random element from subset $S$ and function randPerm($S$) returns a random permutation of $S$.}. 
The order for flipping the decision variables is defined by random vector $R$.  

\begin{algorithm}[h]
\caption{LS($\mathbf{x}$)}
\label{alg:LS}
\begin{algorithmic}[1]
    \STATE $R$ = randPerm($\{1,\ldots,N\}$) 
    \STATE $k=1$ 
    \WHILE{stop criterion is not satisfied}
        \STATE $g=R_k$ 
        \STATE $k=(k \mod N)+1$ 
        \IF {$\delta_g(\mathbf{x}) > 0$}
            \STATE $\mathbf{x} = \mathbf{x}\oplus\mathbf{1}_g$ 
        \ENDIF 
    \ENDWHILE
\end{algorithmic}
\end{algorithm}

Any  pseudo-Boolean function $f: \mathbb{B}^N \rightarrow \mathbb{R}$ can be represented in the following polynomial form \cite{heckendorn2002}:
\begin{equation}
\label{eq:walsh-decomposition}
f(\mathbf{x}) = \sum_{i=0}^{2^N-1} w_i \psi_i(\mathbf{x}) ,
\end{equation}
where $w_i$ is the $i$-th Walsh coefficient, $\psi_i(\mathbf{x}) = (-1)^{\mathbf{i}^\mathrm{T}\mathbf{x}}$ generates a sign, and $\mathbf{i} \in \mathbb{B}^N$ is the binary representation of index $i$. 
Walsh coefficients can be used for identifying nonlinear interactions between variables. 
The interaction between variables is defined as follows. 

\begin{definition}
\label{def:interaction}
Given a pseudo-Boolean function $f: \mathbb{B}^N \rightarrow \mathbb{R}$, we say that variables $x_g$ and $x_h$ \emph{interact} in $f$ if there exists at least one nonzero Walsh coefficient $w_i$ in the Walsh decomposition of $f$ such that the $g$-th and $h$-th elements of $\mathbf{i}$ are equal to one. The \emph{variable interaction} relationship is symmetric.
\end{definition}

For $k$-bounded pseudo-Boolean functions, the Walsh decomposition must be polynomial in size when $k$ is $O(1)$ and $N$ is large because almost all Walsh coefficients are zero. 
Interactions of decision variables in $f$ can be represented by the VIG \cite{chicano2014,tinos2015}. 
The VIG is defined as follows. 

\begin{definition}
\label{def:VIG}
The \emph{variable interaction graph} (VIG) is an undirected graph $G = (V_G,E_G)$, where each vertex $v_i \in V_G$ is related to a decision variable $x_i$, and each edge $(g,h) \in E_G$ indicates that variables $x_g$ and $x_h$ interact in $f$ (Definition~\ref{def:interaction}).
\end{definition}

\section{Local Search with Linkage Learning 2}
\label{sec:LSwLL}
In DLED, when a variable $x_h$ is perturbed in a solution $\mathbf{x}$, the dependencies between $x_h$ and other variables are searched. 
Consider that $\mathbf{y}$ is the solution generated by flipping $x_h$ in $\mathbf{x}$.
Now, consider that a variable $x_g$, $g \neq h$, is flipped in $\mathbf{x}$ and $\mathbf{y}$. 
If flipping $x_g$ results in an improvement in one solution, e.g., $\mathbf{x}$, but not in the other, e.g., $\mathbf{y}$, then variables $x_g$ and $x_h$ interact. 
DLED never returns a false linkage, i.e., an edge not present in the VIG~\cite{przewozniczek2021}. 
However, it can miss some linkages (the same is true for the method proposed in this section). 
In other words, DLED returns an \emph{empirical VIG} that is a partial or complete VIG. 
A disadvantage is that the DLED-based linkage discovery (the construction of an empirical VIG) requires performing a relatively high number of additional solution evaluations.

\citet{tinos2022} used some ideas of DLED in LSwLL to build an empirical VIG during the optimization performed by ILS. 
There are two main differences between LSwLL and DLED. 
First, unlike DLED, the strategy does not need additional fitness evaluations for building the empirical VIG. 
Second, instead of checking improvements as DLED does, the strategy checks the difference in the fitness of the solutions with flipped variables. 
Thus, it is able to find a linkage between two variables even when flipping both results in improving moves. 

Given a candidate solution $\mathbf{x}$, let us define:
\begin{equation} 
	\label{eq:omega}
		\omega_{g,h} (\mathbf{x}) = | \delta_{g}(\mathbf{x}\oplus\mathbf{1}_h)- \delta_g(\mathbf{x}) |
\end{equation}
where $\delta_g(\mathbf{x})$ and $\delta_{g}(\mathbf{x}\oplus\mathbf{1}_h)$ are respectively given by equations \eqref{eq:deltaf3} and \eqref{eq:deltaf4}.  
\citet{tinos2022} show that if $\omega_{g,h} (\mathbf{x}) \neq 0$ for any candidate solution $\mathbf{x} \in \mathbb{B}^N$, then variables $x_h$ and $x_g$ of $f$ interact (according to Definition~\ref{def:interaction}) and $(h,g)$ is an edge of the VIG. 
Thus, while optimizing a candidate solution, LSwLL flips decision variables of the candidate solution in a specific order to detect edges of the VIG. 

Based on LSwLL, we propose LSwLL2, that builds an empirical weighted VIG instead of an empirical VIG. We propose using Eq.~\eqref{eq:omega} to define the strength of the interaction between variables. 

\begin{definition}
\label{def:strenght}
Given a pseudo-Boolean function $f: \mathbb{B}^N \rightarrow \mathbb{R}$, the \emph{strength} of the interaction between variables $x_h$ and $x_g$ in $f$ (Definition~\ref{def:interaction}) is given by:
\begin{equation} 
	\label{eq:upsilon1}
        \upsilon(h,g) = \frac{1}{2^N} \sum_{\mathbf{x} \in \mathbb{B}^N}   \omega_{g,h} (\mathbf{x})		
\end{equation}
where $\omega_{g,h} (\mathbf{x})$ is given by Eq.~\eqref{eq:omega}. 
Strength is symmetric. i.e., $\upsilon(h,g)=\upsilon(g,h).$ 
\end{definition}

\begin{definition}
\label{def:VIGw}
The \emph{weighted variable interaction graph} (VIGw) is an undirected weighted graph $G = (V_G,E_G,W_G)$, where each vertex $v_g \in V_G$ is related to a decision variable $x_g$, and each edge $(h,g) \in E_G$ with nonzero weight $\upsilon(h,g) \in W_G$ indicates that variables $x_h$ and $x_g$ interact in $f$ with strength $\upsilon(h,g)$ by Definition~\ref{def:strenght}.
\end{definition}

Computing $\upsilon(h,g)$ has exponential worst-case time complexity
for arbitrary functions.  Luckily,
the time complexity is polynomial 
for $k$-bounded pseudo-Boolean optimization. 
However, the problem structure must be known \emph{a priori} to do so. 
Here, we will use an \emph{empirical} VIGw instead of the (\emph{true}) VIGw. 

\begin{definition}
\label{def:empiricalVIGw}
Let $G = (V_G,E_G,W_G)$ be the weighted VIG (VIGw) of the problem's instance (Definition~\ref{def:VIGw}). 
An \emph{empirical VIGw}, $G_p = (V_{G_p},E_{G_p},W_{G_p})$, is a graph where $V_{G_p}=V_G$, $E_{G_p} \subset E_G$, and $\hat{\upsilon}(h,g) \in W_{G_p}$ is the weight associated to edge $(h,g)$, given by:
\begin{equation} 
	\label{eq:upsilon2}		
        \hat{\upsilon}(h,g) = \frac{1}{|\Upsilon(h,g)|} \sum_{\mathbf{x} \in \Upsilon(h,g)}  \omega_{g,h} (\mathbf{x}) 
\end{equation}
where $\Upsilon(h,g) \in \mathbb{B}^N$ is the  subset of candidate solutions visited in order to compute the partial sum of $\omega_{g,h} (\mathbf{x})$. 
\end{definition}

The proposed linkage learning strategy is based on the following theorem. 

\begin{theorem}
\label{theorem:LSwLL}
Given a pseudo-Boolean function $f: \mathbb{B}^N \rightarrow \mathbb{R}$, if:
\begin{equation}
    \label{eq:delta_LSwLL}
        \hat{\upsilon}(h,g) > 0
\end{equation}
then variables $x_g$ and $x_h$ interact in $f$ (Definition~\ref{def:interaction}) and $(h,g)$ is an edge of the weighted VIG (VIGw). 
\end{theorem}
\vspace*{-0.3cm}
\begin{proof}
The proof is based on the Walsh representation of $f: \mathbb{B}^N \rightarrow \mathbb{R}$. 
Given a solution $\mathbf{x}$, let us re-write Eq.~\eqref{eq:walsh-decomposition} as follows:
\begin{equation} 
	\label{eq:theor_delta_1}
        f(\mathbf{x}) = \sum_{i \in C} f_i(\mathbf{x})
\end{equation} 
where $f_i(\mathbf{x})=w_i \psi_i(\mathbf{x})$ and $C \subset \{0,1,\ldots,2^N-1\}$ contains only indices of components with nonzero Walsh coefficients. 
Consider the following decomposition of Eq.~\eqref{eq:theor_delta_1}:
\begin{equation*}
\label{eq:theor_delta_2}
    f(\mathbf{x}) = \sum_{\substack{i \in C_{h,g}}} f_i(\mathbf{x}) + 
                    \sum_{\substack{i \in C_g-C_{h,g}}} f_i(\mathbf{x}) +
                    \sum_{\substack{i \in C_h-C_{h,g}}} f_i(\mathbf{x}) + R(\mathbf{x})
\end{equation*}
where: $C_g$ is the subset indicating all components of the sum given by Eq.~\eqref{eq:theor_delta_1} that depend on variable $x_g$; 
$C_h$ is the subset indicating all components that depend on variable $x_h$; and 
$C_{h,g}=C_g \cap C_h$ is the subset indicating all components that depend on both variables $x_g$ and $x_h$.   Finally:
\[ R(\mathbf{x}) =  \sum_{\substack{i \in C-(C_g \cup C_h)}} f_i(\mathbf{x}) \]
sums over all of the remaining Walsh coefficients that do not interact with $g$ or $h$. $R(\mathbf{x})$ does not change when the variables $x_g$ or $x_h$ change.  We can largely ignore $R(\mathbf{x})$ since it does not change.

When variable $x_g$ is flipped, we have:
\begin{equation*}
\label{eq:theor_delta_4}
     f(\mathbf{x}\oplus\mathbf{1}_g) = 
                    \sum_{\substack{i \in C_{h,g}}}  f_i(\mathbf{x}\oplus\mathbf{1}_g) + 
                    \sum_{\substack{i \in C_g-C_{h,g}}}  f_i(\mathbf{x}\oplus\mathbf{1}_g) + 
                    \sum_{\substack{i \in C_h-C_{h,g}}} f_i(\mathbf{x}) + R(\mathbf{x})
\end{equation*}
and when variable $x_h$ is flipped:
\begin{equation*}
\label{eq:theor_delta_5}
    f(\mathbf{x}\oplus\mathbf{1}_h) = 
                    \sum_{\substack{i \in C_{h,g}}} f_i(\mathbf{x}\oplus\mathbf{1}_h) + 
                    \sum_{\substack{i \in C_g-C_{h,g}}} f_i(\mathbf{x}) + 
                    \sum_{\substack{i \in C_h-C_{h,g}}}  f_i(\mathbf{x}\oplus\mathbf{1}_h) + R(\mathbf{x}).
\end{equation*}
If both variables $x_g$ and $x_h$ are flipped, then:
\begin{equation*}
\label{eq:theor_delta_6}
   f(\mathbf{x}\oplus(\mathbf{1}_h+\mathbf{1}_g)) = 
                    \sum_{\substack{i \in \\C_{h,g}}} f_i(\mathbf{x}\oplus(\mathbf{1}_h+\mathbf{1}_g)) + 
                    \sum_{\substack{i \in C_g-\\C_{h,g}}}  f_i(\mathbf{x}\oplus\mathbf{1}_g) 
                    + 
                    \sum_{\substack{i \in C_h-\\C_{h,g}}}  f_i(\mathbf{x}\oplus\mathbf{1}_h) + R(\mathbf{x}). 
\end{equation*}
Thus, from equations \eqref{eq:omega}, \eqref{eq:deltaf3}, \eqref{eq:deltaf4}:
\begin{align*}
\omega_{h,g} (\mathbf{x}) & = | \delta_{g}(\mathbf{x}\oplus\mathbf{1}_h)- \delta_g(\mathbf{x}) |  =
   \big| \big(f(\mathbf{x}\oplus(\mathbf{1}_h+\mathbf{1}_g))- f(\mathbf{x}\oplus\mathbf{1}_h)\big) -
    \big( f(\mathbf{x}\oplus \mathbf{1}_g)-f(\mathbf{x})\big) \big| \\
   & = \sum_{\substack{i \in C_{h,g}}} \big| f_i(\mathbf{x}\oplus(\mathbf{1}_h+\mathbf{1}_g))-
                     f_i(\mathbf{x}\oplus \mathbf{1}_h)-
                     f_i(\mathbf{x}\oplus \mathbf{1}_g)+
                     f_i(\mathbf{x}) \big|
\end{align*}
that is different from zero only if $C_{h,g} \neq \emptyset$. 
As a consequence, $\upsilon(h,g)$ (Eq.~\ref{eq:upsilon1}) and $\hat{\upsilon}(h,g)$ (Eq.~\ref{eq:upsilon2}) are different from zero only when $C_{h,g} \neq \emptyset$.
In other words, if Eq.~\eqref{eq:delta_LSwLL} holds, then variables $x_g$ and $x_h$ interact in $f$, and there exists at least one nonzero Walsh coefficient $w_i$ in the Walsh decomposition of $f$ such that the $g$-th and $h$-th elements of $\mathbf{i}$ are both equal to one (Definition~\ref{def:interaction}).
\end{proof}

\begin{figure}[!ht]
  \centering
  \includegraphics[width=\linewidth]{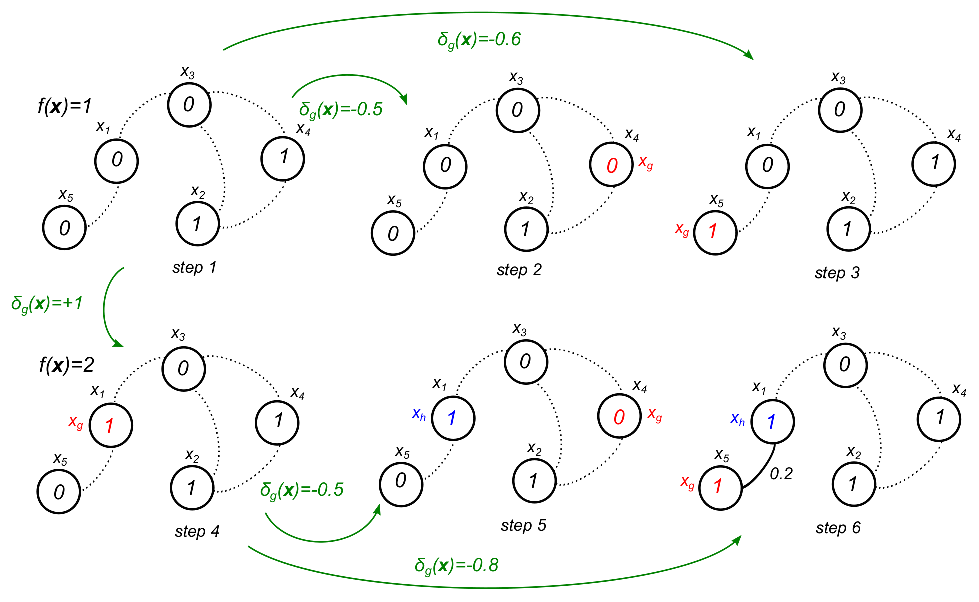}
  \caption{Empirical VIGw in six steps of LSwLL2. 
  In each step, fitness difference resulted from flipping variable $x_g$ is computed. 
  Flipping the variable is accepted only when fitness improvement is detected, i.e., when $\delta_g(\mathbf{x})>0$. In this case, the variable is indicated in the figure by $x_h$.  Initially, the empirical VIGw contains no edges (step 1). Respectively in steps 2 and 3, decision variables $x_4$ and $x_5$ are visited (and stored in list $Q$), resulting in fitness differences equal to -0.5 and -0.6 (stored in list $F$). 
  In step 4, visiting $x_1$ results in fitness improvement. 
  Then, the same sequence of variables (stored in list $Q$) are visited again and the fitness differences are compared to the respective values stored in $F$. In step 6, the fitness difference (-0.8) is not equal to the value stored in $F$ (-0.6), indicating an interaction (edge in the VIGw) between $x_1$ and $x_5$.}
  \label{fig:LSwLL}
\end{figure}

LSwLL2 builds an empirical VIGw (Definition~\ref{def:empiricalVIGw}) by using Eq.~\ref{eq:upsilon2}. 
LSwLL2 is a first-improvement local search used in ILS that optimizes solution $\mathbf{x}$ while updates the empirical VIGw $G_p$. 
When an improvement is detected, local search visits the same sequence of decision variables that were previously flipped. 
The main difference between standard LS and LSwLL2 is that the order for flipping the bits is different. 
The same visiting strategy was used before in LSwLL. 
Figure~\ref{fig:LSwLL} illustrates 6 steps (iterations) of LSwLL2.

Algorithm~\ref{alg:LSwLL} shows the pseudo-code of LSwLL2. 
A list $Q$ is used to store variables that were visited after an improvement (line 27), while a list $F$ is used to store the respective values of $\delta_g(\mathbf{x})$ (line 28). When variables are revisited, the condition given by Eq.~\eqref{eq:delta_LSwLL} is tested by computing  $\delta_g(\mathbf{x}\oplus\mathbf{1}_h)$ and comparing it to the value of $\delta_{g}(\mathbf{x})$ stored in list $F$ (line 6). 
List $R$ (line 1) is used in standard LS for randomly defining the order of the bits that are flipped (Algorithm~\ref{alg:LS}). 
In each iteration of LSwLL2, fitness improvement resulted from flipping $x_g$ in solution $\mathbf{x}$ is checked (line 17). 
If $\delta_g(\mathbf{x})>0$, then $x_g$ is flipped (line 18). 
After revisiting the sequence of variables stored in list $Q$ (line 5), then variables indicated by list $R$ are flipped (line 14). 
A flag ($r$) is used for alternating sequences of iterations where variables are revisited or not. 
In linkage learning, we want to find if variable $x_h$, that resulted in the last improvement (line 19), interacts with variable $x_g$, whose flipping is tested in the current iteration. 
If condition given by Eq.~\eqref{eq:delta_LSwLL} holds, then edge $(h,g)$ is added (if it was not added before) to $G_p$ and its weight is updated (lines 7-11). 
The search stops (line 3) when no improvement is found, i.e., when a local optimum is reached. 

\begin{corollary}
\label{corollary:LSwLL}
LSwLL2 creates an empirical VIGw (Definition~\ref{def:empiricalVIGw}). 
\end{corollary}
\vspace*{-0.3cm}
\begin{proof}
We need to compare the graph $G_p = (V_{G_p},E_{G_p},W_{G_p})$ created by LSwLL2 with the VIGw $G = (V_G,E_G,W_G)$ for the problem's instance. 
We know the number of decision variables ($N$).  
It follows that $V_{G_p}=V_G=\{v_1, v_2, \ldots, v_N\}$, that is one condition for the definition of an empirical VIGw (Definition~\ref{def:empiricalVIGw}). 
Edges ($h,g$) are added by LSwLL2 to $E_{G_p}$, which is initially empty. 
The condition given by Eq.~\eqref{eq:delta_LSwLL} is tested in LSwLL2 only after flipping $x_h$ in $\mathbf{x}$ and computing $f(\mathbf{x}\oplus 1_g)$ and $f(\mathbf{x}\oplus (1_h + 1_g))$. 
Thus, as a consequence of Theorem~\ref{theorem:LSwLL}, an edge $(h,g) \in E_{G_p}$ found by LSwLL2 is also an edge of $E_{G}$, i.e., $E_{G_p}\subset E_{G}$, that is the second condition in Definition~\ref{def:empiricalVIGw}. 
It follows that LSwLL2 never returns a false linkage. 
Finally, $\hat{\upsilon}(h,g)$ is computed in Algorithm~\ref{alg:LSwLL} by using Eq.~\eqref{eq:upsilon2}, which is the third condition in Definition~\ref{def:empiricalVIGw}. 
In Algorithm~\ref{alg:LSwLL}, $\Upsilon(h,g) \in \mathbb{B}^N$ is the subset of visited candidate solutions where the condition given by Eq.~\ref{eq:delta_LSwLL} is true. 
\end{proof}

\begin{algorithm}[!ht]
\caption{LSwLL2($\mathbf{x},G_p$)}
\label{alg:LSwLL}
\begin{algorithmic}[1]
    \STATE $R$=randPerm($\{1,\ldots,N\}$)
    \STATE $j=1$, $k=1$, $r=false$, $Q=\emptyset$, $F=\emptyset$
    \WHILE{stop criterion is not satisfied}
    	\IF {$j < |Q|$} 
            \STATE $g=Q_j$ 
            \IF{ $|\delta_g(\mathbf{x})-F_j|$>0}
                \IF {edge $(h,g) \in E_{G_p}$ }
                    \STATE update weight of edge $(h,g)$ by using Eq.~\eqref{eq:upsilon2}
                \ELSE
                    \STATE add edge $(h,g)$, with weight $|\delta_g(\mathbf{x})-F_j|$, to $G_p$
                \ENDIF
            \ENDIF
        \ELSE
            \STATE $g=R_k$ 
            \STATE $k=(k \mod N)+1$ 
        \ENDIF  
        \IF {$\delta_g(\mathbf{x}) > 0$}
            \STATE $\mathbf{x} = \mathbf{x}\oplus\mathbf{1}_g$ 
            \STATE $h=g$
            \IF {$j <|Q|$ \OR $r==true$}
                \STATE $Q=\emptyset$, $F=\emptyset$ 
            \ENDIF
            \STATE $r = \lnot r$ 
            \STATE j=1
        \ELSE
            \IF {$j \geq |Q|$}
                \STATE add $g$ to list $Q$ 
                \STATE add $\delta_g(\mathbf{x})$ to list $F$ 
            \ENDIF
             \STATE j=j+1
        \ENDIF 
    \ENDWHILE
\end{algorithmic}
\end{algorithm}

It is important to observe that detecting the interactions between two variables using equations \eqref{eq:upsilon2} and \eqref{eq:delta_LSwLL} is similar to what is done by \citet{heckendorn2004}. 
In fact, $\hat{\upsilon}(h,g)$ in Eq.~\eqref{eq:upsilon2} is a \emph{probe}, as defined in ~\cite{heckendorn2004}. 
The equation for probes is even more general than Eq.~\eqref{eq:upsilon2} because probes can also be used to discover linkages for groups with more than two decision variables. 
\citet{heckendorn2004} also suggest that it is possible to use probes to measure the interaction strength between variables, but no algorithms or examples are presented. 
However, their method for detecting variable interactions may require performing a relatively high number of additional solution evaluations for detecting the interaction between two variables. 
On the other hand, LSwLL2 does not need additional solution evaluations because it discovers the empirical VIGw as a side-effect of local search.

\section{VIGw-based Perturbation}
\label{sec:VIGwbP}
Defining an optimal or suboptimal perturbation strategy is not trivial. 
If the perturbation is too strong, ILS often becomes similar to a multi-trial algorithm with random restarts, i.e., there is little correlation between two consecutive local optima generated by ILS.
On the other hand, if the perturbation is too weak, then frequently the two consecutive local optima are equal because the perturbed solution is located at the same basin of attraction of the current local optimum. 
In general, consecutive local optima generated by ILS should be close~\cite{brandao2020}. 
In addition, the evaluation of close local optima is similar in many combinatorial optimization problems.
In this way, jumping between distant local optima often generates a new optimum with less similarity to the current local optimum than when the jumps are between close local optima~\cite{brandao2020}.  

The standard approach for perturbing a solution in ILS is to randomly change $\alpha$ decision variables of the current local optimum. 
When the domain of the candidate solutions is formed by binary strings, this strategy flips a randomly selected subset with $\alpha$ bits.
Clearly setting good values for $\alpha$ depends on fitness landscape properties. 
Adaptive approaches were proposed for avoiding too weak or too strong perturbations~\cite{dowsland2012,hansen2003}. 
Strategies were also proposed for selecting subsets of variables to be changed based on the visited solutions or the importance of decision variables, instead of doing so by using uniform distribution~\cite{battiti1997,brandao2020,lu2009}. 
 
Ideally, a perturbation should maximize the number of flipped variables (distance between solutions) but also minimize the fitness difference between the old and new solutions. 
One strategy for doing this is to pick the best of a subset of random solutions. 
However, this requires evaluating solutions in this subset and, as a consequence, increasing the number of fitness evaluations. 
Alternatively, \citet{tinos2022} propose to use the number of changes that occur in the nonzero Walsh coefficients in Eq.~\eqref{eq:walsh-decomposition} to estimate the fitness difference. 
The \emph{VIG-based perturbation} (VIGbP) basically changes a random decision variable and its neighbors in the VIG. 
The main disadvantage of VIGbP is that when the VIG is too dense, i.e., there are many variables interacting with the random variable, then VIGbP behaves like the standard random perturbation approach, where $\alpha$ random variables are flipped. 

To avoid this problem, we propose here the \emph{VIGw based perturbation} (VIGwbP). 
If the VIGw is not dense, VIGwbP is similar to VIGbP. 
However, if there are many variables interacting with the chosen random variable, then the weights in the VIGw are used to select the neighbors with strongest connections to the random variable. 
Algorithm~\ref{alg:VIGwbP} shows the pseudo-code of VIGwbP. 
A solution $\mathbf{x}$ and a empirical VIGw $G_p$ are the inputs, and a perturbed solution $\mathbf{y}$ is the output. 

\begin{figure}[h]
  \centering
  \includegraphics[width=\linewidth]{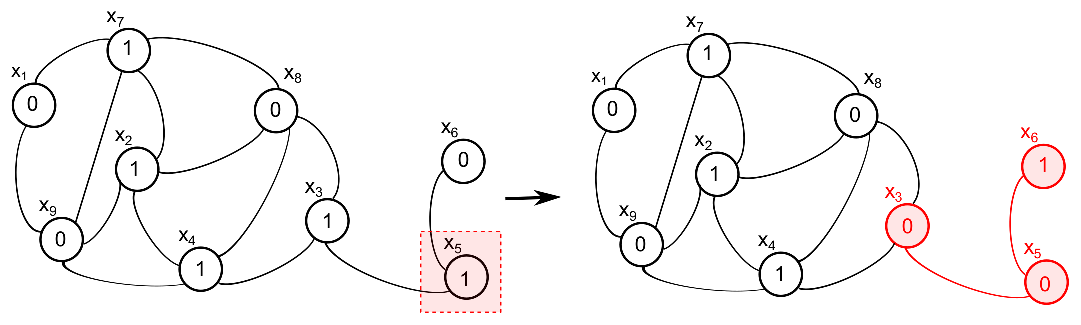}
  \caption{Example of VIGwbP . Decision variable $x_5$ is randomly chosen. Then, $x_5$ and its two neighbors ($x_3$ and $x_6$) in the empirical VIGw (the weights are not showed) are flipped.}
  \label{fig:VIGwbP}
\end{figure}

Let $\Omega_{G_p}(v_i)$ be the subset of vertices with edges incident in vertex $v_i \in V_{G_p}$, i.e., the neighbors of $v_i$ in $G_p$. 
In VIGwbP, a random decision variable $x_i$ (line 3) and 
variables in a subset $L$  are flipped (lines 17-19). 
Subset $L$ can be selected in two different ways:  
\begin{enumerate}[label=\roman*.]
    \item  If $\Omega_{G_p}(v_i)=\emptyset$, i.e., no variable interact with $x_i$ according to the VIGw, then one additional randomly chosen variable $x_j \neq x_i$ (line 15) is flipped. This is done to ensure that at least two bits are changed by perturbation. When vertex $v_i$ has no edges, VIGwbP becomes equal to the standard random perturbation strategy where two random variables are flipped; 
    \item If $\Omega_{G_p}(v_i) \neq \emptyset$, then the variables with the strongest connections to $v_i$ (lines 6-13), according to the VIGw, are flipped. First, variables in $\Omega_{G_p}(v_i)$ are sorted in ascending order according to their weights to $v_i$  (line 6). All variables in $\Omega_{G_p}(v_i)$ with weights to $v_i$ higher than a threshold $\beta$ are added to $L$ (lines 8-13). The threshold $\beta$ is computed by function $thresholdComputation(C)$ in line 7. Function  $thresholdComputation(C)$ creates a sample with the weights of edges $(v_i,C_j)$ for all vertices in $C$ and finds the outliers in this sample by using the box-plot procedure described in~\cite{williamson1989}. The threshold $\beta$ is given by the upper bound outlier; 
\end{enumerate}
It is important to observe that VIGwbP is parameterless. 
Unlike the standard random perturbation where the number of flipped variables is fixed (given by parameter $\alpha$), the number of flipped variables in VIGwbP depends on the weights of the empirical VIGw and on the variables that interact with variable represented by vertex $v_i$. 
When the number of neighbors of $v_i$ is small, e.g., in $k$-bounded pseudo-Boolean functions with low epistasis degree $k$, then all neighbors of $v_i$ in $G_p$ are changed because most of the weights associated to $v_i$ are zero. 
Figure~\ref{fig:VIGwbP} shows an example of VIGwbP where all neighbors of $v_i$ are flipped. 
However, when the number of neighbors of $v_i$ is not small, e.g., when the VIGw is dense, then only the neighbors with strong connections to $v_i$ are flipped. 
Algorithm~\ref{alg:ILS_VIGwbP} shows the pseudo-code of ILS with VIGwbP and LSwLL2.

\begin{algorithm}[ht]
\caption{$\mathbf{y}=$VIGwbP($\mathbf{x},G_p$)}
\label{alg:VIGwbP}
\begin{algorithmic}[1]
    \STATE $\mathbf{y}=\mathbf{x}$, $L=\emptyset$
    \STATE $i$=rand($\{1,\ldots,N\}$)
    \STATE $\mathbf{y} = \mathbf{y}\oplus\mathbf{1}_i$ 
    \STATE $n = |\Omega_{G_p}(v_i)|$ 
    \IF {$n>0$}
        \STATE $C$=sort vertices of $\Omega_{G_p}(v_i)$ in ascending order according to their weights to $v_i$
        \STATE $\beta$=thresholdComputation($C$)
        \STATE add vertex $C_n$ to $L$
        \STATE $j=n-1$
        \WHILE{$j>0$ \AND weight of edge ($v_i,C_j$) greater than $\beta$}
            \STATE add vertex $C_j$ to $L$
            \STATE $j=j-1$
        \ENDWHILE
    \ELSE
       \STATE $L$=rand($V_G-v_i$)
    \ENDIF
    \FOR {$v_j \in L$}
        \STATE $\mathbf{y} = \mathbf{y}\oplus\mathbf{1}_j$ 
    \ENDFOR
    \RETURN $\mathbf{y}$
\end{algorithmic}
\end{algorithm}

\begin{algorithm}[ht]
\caption{ILS with VIGwbP and LSwLL2}
\label{alg:ILS_VIGwbP}
\begin{algorithmic}[1]
    \STATE create $G_p$ with $V_{G_p}=\{v_1, v_2,\ldots , v_N\}$, $E_{G_p}=\emptyset$
    \STATE $\mathbf{x}=$GenerateInitialSolution()
    \STATE $LSwLL2(\mathbf{x},G_p)$
    \WHILE{stop criterion is not satisfied}
        \STATE $\mathbf{y}=$VIGwbP($\mathbf{x},G_p$)
        \STATE $LSwLL2(\mathbf{y},G_p)$
        \STATE $\mathbf{x}=$AcceptanceCriterion($\mathbf{y},\mathbf{x}$)
    \ENDWHILE
\end{algorithmic}
\end{algorithm}

\section{Experiments}
\label{sec:exp}
LSwLL2 creates an empirical VIGw during local search by flipping decision variables in a systematic order that is different from the one adopted by standard LS. 
Information about the interaction between variables is obtained as a side effect of local search. 
An important question is: how does the adopted visiting order and the additional operations needed for estimating the VIGw impact the search? 
In Section~\ref{sec:res_ls} we answer this question. We compare LS and LSwLL2 regarding different measures. 
The VIGw produced by LSwLL2 allows us to obtain new insights about the optimization problem and algorithms. 
We investigate this issue in Section~\ref{sec:res_vig}, where empirical VIGws generated by LSwLL2 are analyzed. 
The VIGws also allow to develop new perturbation strategies for ILS. 
In Section~\ref{sec:res_VIGwbp}, we present results for the comparison of VIGwbP with other perturbation strategies. 
First, the experimental design is presented in Section~\ref{sec:exp_des}.

\subsection{Experimental Design}
\label{sec:exp_des}
We run ILS with LS and LSwLL2 adopting different perturbation strategies. Two different stopping criteria were considered in the experiments: i) fixed number of iterations for each run; ii) fixed runtime. 
ILS is considered in a black box setting, i.e. independent of specific problems.
In the experiments, for all problems, a deterministic acceptance criterion is used by ILS, where a new local optimum $\mathbf{y}$ replaces the current local optimum $\mathbf{x}$ whenever $f(\mathbf{y})>f(\mathbf{x})$. 
Different measures, computed for each run of ILS, were used when comparing algorithms:
\begin{itemize}
    \item \textbf{FIT}: fitness of the best solution found by ILS. In the experiments with 0-1 knapsack, where the fitness of the global optimum is known, we instead use the normalized error \textbf{ERR} (see Section~\ref{sec:expdes_KP}); 
    \item \textbf{PELO}: percentage of escapes from local optima in ILS;    
   \item \textbf{HDLO}: Hamming distance between two consecutive local optima found by ILS;    
    \item \textbf{HDP}: Hamming distance between solutions before and after perturbation;  
    \item \textbf{FDP}: fitness difference for solutions before and after perturbation;    
    \item \textbf{FHRP}: ratio between $FDP$ and $HDP$;
    \item \textbf{NILS}: number of iterations of local search (LS or LSwLL2);   
    \item \textbf{TIME}: runtime in seconds of ILS (for experiments with fixed number of iterations); 
    \item \textbf{NI}: number of iterations of ILS (for experiments with fixed time). 
\end{itemize}
When comparing the performance of different algorithms, the results of the quality ($FIT$ or $ERR$) of the best solution found by ILS are compared in experiments with fixed runtime ($TIME$). 
When analyzing the behavior of different algorithms, measures that depend on the visited local optima and their attraction basins (i.e., $PELO$, $HDLO$, $HDP$, $FDP$, $FHRP$, $NILS$) are compared in experiments with fixed number of iterations of ILS ($NI$). 
In each iteration of ILS (see Algorithm~\ref{alg:ILS_VIGwbP}), a local optimum is visited; as a consequence, in an experiment with fixed $NI$, the number of samples (local optima) is the same for the different algorithms. 

Four perturbation strategies were compared: 
\begin{itemize}
    \item \textbf{SRP, $\alpha=2$}: standard random perturbation with fixed strength $\alpha=2$; 
    \item \textbf{SRP, $\alpha=50$}: standard random perturbation with fixed strength $\alpha=\min \Big(50, \left \lfloor\frac{N}{2} \right\rfloor \Big)$; 
    \item  \textbf{ADP}: adaptive perturbation. ADP is similar to SRP, but $\alpha$ is adapted online. Every 5 iterations of ILS, $\alpha$ is incremented if: i) the perturbation generated a solution in the same basin of attraction of the current local optimum, or ii) the jump caused by perturbation is smaller than the average distance between two consecutive local optima. Otherwise, $\alpha$ is decremented, except for the case where the new local optimum is better than the current local optimum; in this case, $\alpha$ is not changed. The initial (and minimum) value of $\alpha$ is 2, while the maximum value of $\alpha$ is $\left \lfloor\frac{N}{2} \right\rfloor$.
    \item  \textbf{VIGwbP}: VIGw-based perturbation (see Section~\ref{sec:VIGwbP}). 
\end{itemize}
The three first perturbation strategies were run in ILS with LS and in ILS with LSwLL2 for the comparison of LS and LSwLL2. 
The Wilcoxon signed rank test with significance level $0.01$ was used to statistically compare the results of an algorithm with LS to the results of the same algorithm with LSwLL2. 
ILS with LSwLL2 was also used for the comparison of different perturbation strategies. 
When comparing the results of VIGwbP to the other perturbation strategies, the Wilcoxon signed rank test corrected with the Bonferroni-Holm procedure for multiple comparisons was used~\cite{latorre2021}. 
A workstation with processor Intel@ Core i7-12700 Alder Lake (12 cores, 20 threads, 25 MB cache) and 128 GB of RAM was used for running the algorithms implemented in C++\footnote{The source code for VIGwbP and LSwLL2 is freely available on Zenodo at
\url{https://zenodo.org/records/10694142}.
}. 
ILS was applied to three optimization problems (NK landscapes, 0-1 knapsack problem, and feature selection). 
In the three problems, the evaluation function $f(\mathbf{x})$ must be maximized. 

\subsubsection{NK Landscapes}
NK landscapes are pseudo-Boolean functions defined as:
\begin{equation} 
	\label{eq:eval_NK}
        f(\mathbf{x}) = \sum_{i=1}^{N} f_i(\mathbf{z}_i)
\end{equation} 
where $\mathbf{z}_i \in \mathbb{B}^k$ contains variable $x_{i}$ and other $K \geq 1$ variables of $\mathbf{x}$, i.e., the number of inputs for each subfunction $f_i$ is always $k=K+1$. 
The values of $f_i$ are randomly generated for all the assignments of $\mathbf{z}_i$. 
There are different ways of selecting the $K$ other variables in $\mathbf{z}_i$. 
The most popular models are the adjacent and random models. 
In the adjacent model~\cite{wright2000}, each subfunction $f_i$ depends on variable $x_{i}$ and $K$ other adjacent variables ($x_{i+1}$, $x_{i+2}$, \ldots). 
In the random model, each subfunction $f_i$ depends on variable $x_{i}$ and $K$ other random variables. 

Results of experiments with adjacent and random NK landscapes for $N=\{100, 500, 1000\}$ and $k=\{3, 5\}$ are presented. 
Five instances were generated for each model and combination of $N$ and $k$. 
The number of runs for each instance was ten. 
Thus, there were 50 runs generated for each model and combination of $N$ and $k$.   
In experiments with fixed number of iterations, the number of iterations of ILS in each run was $NI=5000$. 
In experiments with fixed time, the runtime was $TIME=\frac{N k}{5}$ seconds. 
These limits (and those for the other two problems) were obtained in initial experiments of ILS with LS and SRP, not shown here. 

\subsubsection{0-1 Knapsack Problem}
\label{sec:expdes_KP}
In the knapsack problem, the most profitable items must be selected, given a limited knapsack capacity. 
There are different versions of the knapsack problem. 
Here, the 0-1 knapsack problem described in~\cite{han2000} is considered where the objective function is defined as:
\begin{equation}
\label{eq:knap_f}
f(\mathbf{x})=\sum_{i=1}^{N} {p_i x_i} - r(\mathbf{x})
\end{equation}
where $\mathbf{x} \in \mathbb{B}^N$ represents a subset of items in the knapsack. 
The $i$-th item has weight $w_i \in \mathbb{R}^{+}$ and profit $p_i \in \mathbb{R}^{+}$. 
Here, the weights are randomly generated in the range $[5,20]$ and the profits are randomly generated in the range $[40, 100]$. 
The knapsack capacity $C$ is equal to 50$\%$ of the sum of all weights. 
The penalty term $r(\mathbf{x})$ is given by:
\begin{equation}
\label{eq:knap_p}
 r(\mathbf{x})= \left\{\!
\begin{array}{ll}
0, & \textrm{if $ \sum_{i=1}^{N} {w_i x_i} \leq C $} \\
\Big( \sum_{i=1}^{N} {w_i x_i} - C \Big) \max_{i=1,\ldots,N} (p_i/w_i), & \textrm{otherwise}
\end{array} \right.
\end{equation}
Results of experiments for $N=\{500, 1000, 1500, 2000\}$ are presented. 
Five instances were generated for each $N$. 
The number of runs for each instance was 10. 
Thus, there were 50 runs generated for each $N$.
In experiments with fixed number of iterations, the number of iterations of ILS in each run was $NI=30000$. 
In experiments with fixed time, the runtime was $TIME=\frac{N}{2}$ seconds. 
In the tables with experimental results, we show the normalized error ($ERR$), i.e., $\frac{f(\mathbf{x}^*)-f(\mathbf{x})}{f(\mathbf{x}^*)}$, where $f(\mathbf{x}^*)$ and $f(\mathbf{x})$ are respectively the evaluation of the global optimum and the evaluation of the best solution found by ILS. 
We used dynamic programming for finding the global optimum. 
The time complexity of dynamic programming for the 0-1 knapsack problem is $O(N C)$.
If $C$ is polynomial in $N$, dynamic programming runs in polynomial time; however, the problem is NP-hard for the general case~\cite{andonov2000}.

\subsubsection{Feature Selection}
Feature selection is an important task in data mining~\cite{xue2016}. 
In the wrapper approach, an optimization algorithm is used for finding a subset of features specific to a dataset and a machine learning model. 
For pseudo-Boolean optimization algorithms, a candidate solution $\mathbf{x} \in \mathbb{B}^N$ indicates a subset of selected features . 
Here, the machine learning model is the \emph{K-nearest neighbors} (KNN) algorithm, with $K=3$. 
We applied ILS for feature selection for classification and regression datasets. 
The evaluation function has two terms:
\begin{equation}
\label{eq:fs_f}
f(\mathbf{x})= 0.98 f_{1}(\mathbf{x}) + 0.02 \frac{N- \sum_{i=1}^{N} {x_i}}{N}
\end{equation} 
where $f_{1}(\mathbf{x})$ is a measure for the performance of the machine learning model and the second term depends on the number of selected features.  
In classification, $f_{1}(\mathbf{x})$ is the rate of examples in the test set correctly classified by the machine learning model with features indicated by $\mathbf{x}$. 
In regression, $f_{1}(\mathbf{x})=1-E(\mathbf{x})$, where $E(\mathbf{x})$ is the mean squared error for the test set when the machine learning model with features indicated by $\mathbf{x}$ is used.  
Here, the size of the training and test sets were respectively $0.7n_{ex}$ and $0.3n_{ex}$, where $n_{ex}$ is the number of examples (samples) in the dataset. 
 
Results for 8 datasets are presented. 
The datasets, listed in Table~\ref{tab:datasets}, are from the UCI Machine Learning Repository \cite{dua2019}, except for covidxr, cnae9m and friedman.  
Dataset covidxr~\cite{tinos2020} was obtained by extracting radionomic features from chest x-ray images. 
The images were selected at random from the COVIDx training dataset \cite{wang2020}, a public dataset containing x-ray images of three classes: normal, and patients with COVID19 and non-COVID19 pneumonia.  
By using PyRadiomics \cite{van2017}, 93 texture features were extracted from the images.  
Dataset cnae9m was obtained by selecting the first 500 examples of the UCI Machine Learning Repository dataset cnae. 
In addition, 74 out of 856 features of the original dataset were selected. 
The selected features are those with frequency equal or higher than 10, i.e., they appear in at least 10 examples of the dataset. 
Dataset friedman is based on a synthetic data  generator model proposed by \citet{friedman2008}, where the interaction between features in machine learning was investigated. 
The model generates desired outputs that depend on 8 out of 100 random discrete variables, $u_i$. 
The target function is:
\begin{equation}
\label{eq:f_friedman}
F(\mathbf{u})= 9 \prod_{j=0}^{2}e^{-3(1-u_j)^2} - 0.8 e^{-2(u_3-u_4)} + 2 \sin^2(\pi u_5) - 2.5(u_6-u_7)+\epsilon
\end{equation} 
where the noise, $\epsilon \sim \mathcal{N}(0,\,\sigma^{2})$, is generated with standard deviation $\sigma$ chosen to produce a two-to-one signal-to-noise ratio.
Here, $n_{ex}=500$.

The number of runs for each dataset was 25. 
In experiments with fixed number of iterations, the number of iterations of ILS in each run was $NI=1000$. 
In experiments with fixed time, the runtime was $TIME=\frac{N n_{ex}}{10}$ seconds.

\begin{table}[h]
\scriptsize
\centering
\caption{Datasets in the feature selection problem.}
\begin{tabular}{r|rrrr}
  \toprule
dataset & type & features ($N$) & examples ($n_{ex}$) & outputs or classes\\  
\midrule
housing &regression & 13 & 506 &  1 \\
friedman &regression & 100 & 500 &  1 \\ 
ionosphere & classification & 34 & 351 & 2 \\
sonar & classification & 60 & 208 & 2 \\
cnae9m & classification & 74 & 500 & 9 \\
libras & classification & 90 & 360 & 15 \\
covidxr & classification & 93 & 456 & 3 \\
arrhythmia & classification & 279 & 452 & 16 \\  
 \bottomrule
\end{tabular}
\label{tab:datasets}
\end{table}

 \subsection{Results: comparing local search strategies}
 \label{sec:res_ls}
Tables~\ref{tab:res_NK_fixed_gen_1}-\ref{tab:res_FS_fixed_gen} show the results for the experiments with fixed number of iterations. 
The median of different measures are presented, as well as the statistical comparison between the results of algorithms with LS (SRP with $\alpha=2$, SRP with $\alpha=50$, and ADP) and the results of the respective algorithms with LSwLL2. 
Tables S1-S6 in the Supplementary Material show the p-values for the Wilcoxon signed rank test used for this comparison. 
In the Supplementary Material (tables S19-S25), median and best FIT and ERR values for the different instances of NK landscapes and 0-1 knapsack problem are also presented. 
Some observations can be made from the results. 

In all experiments with fixed number of iterations, the null hypothesis (no statistical difference between the results) cannot be rejected when comparing $FIT$ between each ILS with LS and the respective algorithm with LSwLL2. 
For the other measures, with exception for $NILS$ and $TIME$, ILS with LS presented similar results to ILS with LSwLL2. 
For some instances, ILS with LS presented better results than ILS with LSwLL2, and vice-verse, but statistical difference between the results was observed in few cases. 
Similar results were obtained when the results for the different instances are analyzed.  

However, for the number of iterations of local search ($NILS$) and runtime ($TIME$), ILS with LS generally presented statistically better results when compared to ILS with LSwLL2. 
LS was generally faster because revisiting variables implies in higher $NILS$ and because some operations are needed to build and manipulate the empirical VIGw.  
In this way, one can ask what happens when the algorithms runs for the same time. 
If one algorithm is faster than the other, then ILS will run for more iterations, and better solutions can potentially be achieved.  

Tables~\ref{tab:res_NK_fixed_time}-\ref{tab:res_FS_fixed_time} show the results for the experiments with fixed time. 
In these tables, only $FIT$ (or $ERR$) and $NI$ are presented.  Tables S7-S9 in the Supplementary Material show the respective p-values for the statistical comparison.
In fact, when the runtime is fixed, the number of iterations of ILS ($NI$) is higher for the algorithm with LS, when compared to the respective algorithm with LSwLL2. 
However, the results for $FIT$ (or $ERR$) were similar, except for: 2 out of 36 cases (3 algorithms for each of the 12 combinations of model, $N$, and $K$) for NK landscapes; 4 out of 12 cases (3 algorithms for each of the 4 values of $N$) for the 0-1 knapsack problem; 1 out of 24 cases (3 algorithms for each of the 8 datasets) for the feature selection problem. 
Similar results are obtained when the runtime is changed; tables S36-S31 and S32-S37 in the Supplementary Material respectively show the results for half and one-third of the runtime ($TIME$) of the original experiments. 
Similar results are also obtained when a different accepting criterion is used; tables S38-S43 in the Supplementary Material show the results for experiments where a simulated annealing acceptance criterion is used.
Thus, in all the experiments, the adopted visiting order and the additional operations needed for estimating the VIGw in LSwLL2 did not significantly impact the performance of ILS.

\begin{table}[h]
\scriptsize
\centering
\caption{Median of different measures for the NK landscapes experiment with fixed number of iterations.
 The symbols `$=$', `$+$', and `$-$' respectively indicate that the median for the Alg. $A$ with LSwLL2 is equal, better or worse than the median of Alg. $A$ with LS. 
 The Wilcoxon signed rank test was employed to statistically compare the results of Alg. $A$ with LSwLL2 and with LS. 
 The letter $s$ indicates that the null hypothesis (no statistical difference between the results) can be rejected according to the significance level $0.01$.
The best $FIT$ results among the algorithms with LSwLL2 are in bold.
}
\begin{tabular}{r|rrr|rrr|rrrr}
  \toprule
        &  \multicolumn{3}{|c}{} & \multicolumn{3}{|c}{LS} & \multicolumn{4}{|c}{LSwLL2}  \\
measure & model & $N$ & $k$ & SRP, $\alpha=2$ & SRP, $\alpha=50$ & ADP & SRP, $\alpha=2$ & SRP, $\alpha=50$ & ADP & VIGwbP\\  
\midrule
FIT & adjacent &100 &3 &0.7517 &0.7465 &0.7523 &0.7517(=) &0.7466(+) &\textbf{0.7523}(=) &\textbf{0.7523} \\ 
 &  &  &5 &0.7728 &0.7508 &0.7661 &0.7702(-) &0.7478(-) &0.7676(+) &\textbf{0.7765} \\ 
 &  &500 &3 &0.7448 &0.7414 &0.7453 &0.7449(+) &0.7415(+) &0.7455(+) &\textbf{0.7463} \\ 
 &  &  &5 &0.7702 &0.7537 &0.7710 &0.7702(-) &0.7530(-) &0.7697(-) &\textbf{0.7737} \\ 
 &  &1000 &3 &0.7455 &0.7411 &0.7461 &0.7455(-) &0.7410(-) &0.7462(+) &\textbf{0.7467} \\ 
 &  &  &5 &0.7709 &0.7551 &0.7708 &0.7704(-) &0.7556(+) &0.7694(-) &\textbf{0.7761} \\ 
 & random &100 &3 &0.7527 &0.7490 &0.7538 &\textbf{0.7538}(+) &0.7477(-) &\textbf{0.7538}(=) &\textbf{0.7538} \\ 
 &  &  &5 &0.7729 &0.7752 &0.7801 &0.7707(-) &0.7746(-) &\textbf{0.7825}(+) &0.7813 \\ 
 &  &500 &3 &0.7484 &0.7505 &0.7506 &0.7483(-) &0.7496(-) &0.7492(-) &\textbf{0.7507} \\ 
 &  &  &5 &0.7657 &0.7768 &0.7711 &0.7671(+) &\textbf{0.7768}(-) &0.7706(-) &0.7726 \\ 
 &  &1000 &3 &0.7462 &0.7470 &0.7474 &0.7464(+) &0.7468(-) &0.7474(-) &\textbf{0.7476} \\ 
 &  &  &5 &0.7679 &0.7772 &0.7708 &0.7677(-) &\textbf{0.7750}(-) &0.7709(+) &0.7720 \\ \hline
 PELO & adjacent &100 &3 &0.4054 &1.0000 &0.8707 &0.3919(s-) &1.0000(=) &0.8681(-) &0.7745 \\ 
 &  &  &5 &0.5539 &1.0000 &0.9677 &0.5575(+) &1.0000(=) &0.9721(+) &0.9827 \\ 
 &  &500 &3 &0.4224 &1.0000 &0.8663 &0.4191(s-) &1.0000(=) &0.8671(+) &0.7023 \\ 
 &  &  &5 &0.5512 &1.0000 &0.9386 &0.5476(s-) &1.0000(=) &0.9444(+) &0.9504 \\ 
 &  &1000 &3 &0.4131 &1.0000 &0.8673 &0.4075(s-) &1.0000(=) &0.8675(+) &0.6249 \\ 
 &  &  &5 &0.5690 &1.0000 &0.9492 &0.5601(s-) &1.0000(=) &0.9525(+) &0.9155 \\ 
 & random &100 &3 &0.2939 &1.0000 &0.8636 &0.2805(s-) &1.0000(=) &0.8630(-) &0.7222 \\ 
 &  &  &5 &0.2856 &1.0000 &0.9512 &0.2784(-) &1.0000(=) &0.9466(-) &0.8961 \\ 
 &  &500 &3 &0.2806 &0.9998 &0.8616 &0.2625(s-) &0.9998(=) &0.8618(+) &0.6185 \\ 
 &  &  &5 &0.2760 &1.0000 &0.8642 &0.2647(-) &1.0000(=) &0.8646(+) &0.8896 \\ 
 &  &1000 &3 &0.2956 &0.9998 &0.8633 &0.2833(s-) &0.9998(=) &0.8623(-) &0.5803 \\ 
 &  &  &5 &0.2852 &0.9996 &0.8648 &0.2682(s-) &0.9998(+) &0.8639(-) &0.8409 \\ \hline
HDLO & adjacent &100 &3 &2.9943 &32.3701 &5.5185 &2.9971(s-) &32.5352(-) &5.4953(+) &3.7738 \\ 
 &  &  &5 &3.4243 &45.2870 &12.3938 &3.4106(+) &45.3933(-) &12.1533(+) &6.8049 \\ 
 &  &500 &3 &2.9331 &31.0867 &5.4897 &2.9294(s+) &31.0312(+) &5.5008(-) &3.6152 \\ 
 &  &  &5 &3.4644 &53.1628 &8.1556 &3.4791(-) &53.7060(s-) &8.4569(-) &6.4728 \\ 
 &  &1000 &3 &2.9489 &30.4221 &5.5340 &2.9483(s+) &30.3254(+) &5.5381(-) &3.5391 \\ 
 &  &  &5 &3.4890 &53.5861 &9.1734 &3.5157(s-) &53.6753(-) &11.7579(-) &6.1398 \\ 
 & random &100 &3 &3.1472 &35.5251 &6.8367 &3.1884(s-) &35.5701(-) &6.8722(-) &5.6853 \\ 
 &  &  &5 &3.5098 &45.9222 &24.8166 &3.5714(-) &46.3190(-) &22.2663(+) &16.6734 \\ 
 &  &500 &3 &3.1859 &24.3748 &6.4308 &3.2016(s-) &24.1341(+) &6.4847(-) &5.1956 \\ 
 &  &  &5 &3.4478 &38.0247 &7.6396 &3.5010(-) &37.4788(+) &7.6886(-) &12.6422 \\ 
 &  &1000 &3 &3.1940 &23.3178 &6.2990 &3.2140(s-) &23.5647(-) &6.3613(-) &4.8439 \\ 
 &  &  &5 &3.5311 &28.9251 &7.2440 &3.6035(s-) &29.9807(-) &7.3297(-) &10.2714 \\ \hline
 HDP & adjacent &100 &3 &2.0000 &50.0000 &7.5403 &2.0000(=) &50.0000(=) &7.5173(+) &4.8648 \\ 
 &  &  &5 &2.0000 &50.0000 &12.0044 &2.0000(=) &50.0000(=) &11.5605(+) &8.8053 \\ 
 &  &500 &3 &2.0000 &50.0000 &7.7712 &2.0000(=) &50.0000(=) &7.7113(+) &4.3436 \\ 
 &  &  &5 &2.0000 &50.0000 &8.0045 &2.0000(=) &50.0000(=) &8.2698(-) &8.0285 \\ 
 &  &1000 &3 &2.0000 &50.0000 &8.0104 &2.0000(=) &50.0000(=) &8.0024(+) &3.8107 \\ 
 &  &  &5 &2.0000 &50.0000 &8.7610 &2.0000(=) &50.0000(=) &11.2783(-) &7.0442 \\ 
 & random &100 &3 &2.0000 &50.0000 &11.1383 &2.0000(=) &50.0000(=) &11.2383(s-) &6.7105 \\ 
 &  &  &5 &2.0000 &50.0000 &24.4440 &2.0000(=) &50.0000(=) &21.9208(+) &15.7501 \\ 
 &  &500 &3 &2.0000 &50.0000 &12.8740 &2.0000(=) &50.0000(=) &12.9185(-) &6.1961 \\ 
 &  &  &5 &2.0000 &50.0000 &13.5094 &2.0000(=) &50.0000(=) &13.5432(-) &18.7361 \\ 
 &  &1000 &3 &2.0000 &50.0000 &12.3188 &2.0000(=) &50.0000(=) &12.4118(-) &5.4730 \\ 
 &  &  &5 &2.0000 &50.0000 &12.9781 &2.0000(=) &50.0000(=) &13.5448(-) &16.7578 \\ \hline
FDP & adjacent &100 &3 &0.0179 &0.2440 &0.0652 &0.0180(-) &0.2436(+) &0.0648(+) &0.0192 \\ 
 &  &  &5 &0.0285 &0.2443 &0.1269 &0.0283(s+) &0.2434(+) &0.1248(+) &0.0363 \\ 
 &  &500 &3 &0.0035 &0.0765 &0.0134 &0.0035(-) &0.0767(-) &0.0132(+) &0.0037 \\ 
 &  &  &5 &0.0056 &0.1068 &0.0219 &0.0056(+) &0.1067(+) &0.0223(-) &0.0071 \\ 
 &  &1000 &3 &0.0017 &0.0397 &0.0068 &0.0017(+) &0.0398(-) &0.0069(-) &0.0018 \\ 
 &  &  &5 &0.0027 &0.0587 &0.0118 &0.0027(+) &0.0590(-) &0.0148(-) &0.0033 \\ 
 & random &100 &3 &0.0179 &0.2443 &0.0879 &0.0180(-) &0.2440(+) &0.0890(s-) &0.0487 \\ 
 &  &  &5 &0.0289 &0.2713 &0.2033 &0.0292(-) &0.2720(-) &0.1989(+) &0.1637 \\ 
 &  &500 &3 &0.0036 &0.0815 &0.0232 &0.0036(+) &0.0814(+) &0.0228(+) &0.0095 \\ 
 &  &  &5 &0.0058 &0.1219 &0.0379 &0.0058(+) &0.1223(-) &0.0381(-) &0.0483 \\ 
 &  &1000 &3 &0.0018 &0.0422 &0.0109 &0.0018(+) &0.0422(+) &0.0111(-) &0.0042 \\ 
 &  &  &5 &0.0029 &0.0674 &0.0188 &0.0029(-) &0.0669(+) &0.0193(-) &0.0218 \\ 
 \bottomrule
\end{tabular}
\label{tab:res_NK_fixed_gen_1}
\end{table} 
 
\begin{table}[h]
\scriptsize
\centering
\caption{Continuation of Table \ref{tab:res_NK_fixed_gen_1}.}
\begin{tabular}{r|rrr|rrr|rrrr}
  \toprule
        &  \multicolumn{3}{|c}{} & \multicolumn{3}{|c}{LS} & \multicolumn{4}{|c}{LSwLL2}  \\
measure & model & $N$ & $k$ & SRP, $\alpha=2$ & SRP, $\alpha=50$ & ADP & SRP, $\alpha=2$ & SRP, $\alpha=50$ & ADP & VIGwbP\\  
\midrule 
FHRP & adjacent &100 &3 &0.0090 &0.0049 &0.0083 &0.0090(-) &0.0049(+) &0.0083(-) &0.0039 \\ 
 &  &  &5 &0.0142 &0.0049 &0.0106 &0.0141(s+) &0.0049(+) &0.0108(-) &0.0041 \\ 
 &  &500 &3 &0.0017 &0.0015 &0.0017 &0.0017(-) &0.0015(-) &0.0017(-) &0.0009 \\ 
 &  &  &5 &0.0028 &0.0021 &0.0027 &0.0028(+) &0.0021(+) &0.0027(+) &0.0009 \\ 
 &  &1000 &3 &0.0009 &0.0008 &0.0009 &0.0009(+) &0.0008(-) &0.0009(+) &0.0005 \\ 
 &  &  &5 &0.0014 &0.0012 &0.0013 &0.0014(+) &0.0012(-) &0.0013(+) &0.0005 \\ 
 & random &100 &3 &0.0090 &0.0049 &0.0082 &0.0090(-) &0.0049(+) &0.0081(+) &0.0072 \\ 
 &  &  &5 &0.0145 &0.0054 &0.0083 &0.0146(-) &0.0054(-) &0.0086(-) &0.0104 \\ 
 &  &500 &3 &0.0018 &0.0016 &0.0018 &0.0018(+) &0.0016(+) &0.0018(+) &0.0015 \\ 
 &  &  &5 &0.0029 &0.0024 &0.0028 &0.0029(+) &0.0024(-) &0.0028(+) &0.0026 \\ 
 &  &1000 &3 &0.0009 &0.0008 &0.0009 &0.0009(+) &0.0008(+) &0.0009(+) &0.0008 \\ 
 &  &  &5 &0.0014 &0.0013 &0.0014 &0.0014(-) &0.0013(+) &0.0014(-) &0.0013 \\ \hline
NILS & adjacent &100 &3 &186.2 &318.8 &231.9 &206.3(s-) &396.4(s-) &272.8(s-) &219.7 \\ 
 &  &  &5 &194.8 &334.1 &272.8 &209.2(s-) &407.9(s-) &323.6(s-) &234.8 \\ 
 &  &500 &3 &924.7 &1535.2 &1183.9 &1022.3(s-) &1934.2(s-) &1384.8(s-) &1083.0 \\ 
 &  &  &5 &966.0 &1721.8 &1258.2 &1050.8(s-) &2166.2(s-) &1489.0(s-) &1173.5 \\ 
 &  &1000 &3 &1856.8 &3036.9 &2360.0 &2054.2(s-) &3844.1(s-) &2790.6(s-) &2148.3 \\ 
 &  &  &5 &1928.9 &3390.9 &2533.3 &2098.7(s-) &4280.2(s-) &3086.7(s-) &2299.4 \\ 
 & random &100 &3 &189.4 &344.1 &257.3 &212.0(s-) &426.9(s-) &309.5(s-) &276.0 \\ 
 &  &  &5 &203.8 &405.5 &357.7 &226.4(s-) &496.8(s-) &424.5(s-) &408.5 \\ 
 &  &500 &3 &956.0 &1619.5 &1303.1 &1075.1(s-) &2056.7(s-) &1594.5(s-) &1366.1 \\ 
 &  &  &5 &1016.4 &2097.9 &1471.3 &1137.9(s-) &2679.7(s-) &1797.6(s-) &2018.4 \\ 
 &  &1000 &3 &1900.1 &3114.9 &2566.5 &2137.5(s-) &3958.0(s-) &3109.1(s-) &2626.3 \\ 
 &  &  &5 &2038.0 &3797.3 &2837.8 &2281.6(s-) &4880.3(s-) &3513.2(s-) &3763.7 \\ \hline
TIME & adjacent &100 &3 &3.9650 &6.7050 &4.9650 &4.3500(s-) &8.3250(s-) &5.7500(s-) &4.6500 \\ 
 &  &  &5 &6.5300 &11.2600 &9.4800 &7.0250(s-) &13.6000(s-) &10.9200(s-) &8.1150 \\ 
 &  &500 &3 &94.4050 &156.2500 &120.3100 &103.8650(s-) &195.0200(s-) &140.5550(s-) &109.7450 \\ 
 &  &  &5 &156.8600 &264.6500 &198.6450 &169.9450(s-) &333.4500(s-) &227.9350(s-) &180.9650 \\ 
 &  &1000 &3 &370.5050 &572.3600 &459.7000 &405.7100(s-) &721.7900(s-) &525.9000(s-) &402.4600 \\ 
 &  &  &5 &581.8800 &913.7850 &756.6650 &644.7250(s-) &1055.6150(s-) &868.6000(s-) &662.6400 \\ 
 & random &100 &3 &3.7950 &7.0350 &5.3500 &4.2800(s-) &8.5950(s-) &6.4100(s-) &5.5900 \\ 
 &  &  &5 &6.8300 &13.7350 &12.0950 &7.6000(s-) &16.4850(s-) &14.0700(s-) &13.8550 \\ 
 &  &500 &3 &97.3750 &165.2900 &132.7050 &109.1800(s-) &206.3700(s-) &161.4350(s-) &138.4850 \\ 
 &  &  &5 &164.5400 &337.6700 &237.3000 &184.4300(s-) &413.2150(s-) &289.5600(s-) &325.0850 \\ 
 &  &1000 &3 &369.7400 &600.7250 &495.1350 &405.4150(s-) &754.5200(s-) &599.4250(s-) &498.3650 \\ 
 &  &  &5 &633.7000 &1123.5300 &865.7950 &694.9200(s-) &1372.4000(s-) &1048.7600(s-) &1140.4350 \\ 
\bottomrule
\end{tabular}
\label{tab:res_NK_fixed_gen_2}
\end{table} 
 
\begin{table}[h]
\scriptsize
\centering
\caption{Median of different measures for the 0-1 knapsack problem experiment with fixed number of iterations.
}
\begin{tabular}{r|r|rrr|rrrr}
  \toprule
        &  \multicolumn{1}{|c}{} & \multicolumn{3}{|c}{LS} & \multicolumn{4}{|c}{LSwLL2}  \\
measure & $N$ & SRP, $\alpha=2$ & SRP, $\alpha=50$ & ADP & SRP, $\alpha=2$ & SRP, $\alpha=50$ & ADP & VIGwbP\\  
\midrule
ERR &500 &0.0020 &0.1030 &0.1013 &\textbf{0.0020}(+) &0.1042(-) &0.0974(+) &0.0156 \\ 
 &1000 &0.0037 &0.1078 &0.1152 &\textbf{0.0038}(-) &0.1058(+) &0.1170(-) &0.0453 \\ 
 &1500 &0.0054 &0.1100 &0.1240 &\textbf{0.0052}(+) &0.1111(-) &0.1230(+) &0.0634 \\ 
 &2000 &0.0065 &0.1110 &0.1304 &\textbf{0.0065}(+) &0.1122(-) &0.1316(-) &0.0683 \\  \hline
 PELO &500 &0.9981 &1.0000 &1.0000 &0.9980(-) &1.0000(=) &1.0000(=) &0.9999 \\ 
 &1000 &0.9991 &1.0000 &1.0000 &0.9992(+) &1.0000(=) &1.0000(=) &1.0000 \\ 
 &1500 &0.9995 &1.0000 &1.0000 &0.9995(+) &1.0000(=) &1.0000(=) &1.0000 \\ 
 &2000 &0.9997 &1.0000 &1.0000 &0.9997(-) &1.0000(=) &1.0000(=) &1.0000 \\  \hline
 HDLO &500 &3.5023 &55.2482 &191.8409 &3.4488(s+) &55.0236(s+) &189.6850(s+) &47.5625 \\ 
 &1000 &3.5394 &56.5045 &401.4311 &3.5102(s+) &56.2265(s+) &400.5841(+) &118.4125 \\ 
 &1500 &3.5840 &56.9805 &608.2850 &3.5505(s+) &56.7301(s+) &605.8226(+) &118.2869 \\ 
 &2000 &3.6280 &57.2749 &807.1189 &3.5916(s+) &56.9749(s+) &805.2528(+) &112.6752 \\ \hline
 HDP &500 &2.0000 &50.0000 &190.9133 &2.0000(=) &50.0000(=) &188.7388(s+) &39.6193 \\ 
 &1000 &2.0000 &50.0000 &400.1665 &2.0000(=) &50.0000(=) &399.3409(+) &107.7639 \\ 
 &1500 &2.0000 &50.0000 &606.6248 &2.0000(=) &50.0000(=) &604.2022(+) &108.1878 \\ 
 &2000 &2.0000 &50.0000 &805.0636 &2.0000(=) &50.0000(=) &803.2529(+) &103.4867 \\ \hline
 FDP &500 &200.0370 &1392.2869 &4016.7035 &200.3043(-) &1390.9637(+) &3982.2099(+) &2255.5206 \\ 
 &1000 &199.6593 &1388.7759 &7096.4020 &199.7152(-) &1387.0619(+) &7003.8640(+) &4158.8532 \\ 
 &1500 &198.1436 &1386.6128 &9692.1652 &198.2849(-) &1378.9760(+) &9701.8203(-) &3632.7281 \\ 
 &2000 &197.2403 &1373.1543 &12003.5773 &197.4930(-) &1368.1290(+) &11776.6008(+) &3255.7787 \\  \hline
 FHRP &500 &100.0185 &27.8457 &20.9501 &100.1521(-) &27.8193(+) &21.1822(-) &57.0187 \\ 
 &1000 &99.8297 &27.7755 &17.6376 &99.8576(-) &27.7412(+) &17.4459(+) &38.7048 \\ 
 &1500 &99.0718 &27.7323 &15.9433 &99.1424(-) &27.5795(+) &16.0497(-) &33.6937 \\ 
 &2000 &98.6202 &27.4631 &14.8609 &98.7465(-) &27.3626(+) &14.6713(+) &31.4852 \\ \hline
 NILS &500 &581.4 &586.8 &566.7 &578.5(s+) &591.5(s-) &576.8(s-) &641.9 \\ 
 &1000 &1144.2 &1139.9 &1078.7 &1134.2(s+) &1143.8(s-) &1092.1(s-) &1137.7 \\ 
 &1500 &1701.9 &1682.3 &1591.1 &1687.1(s+) &1686.3(s-) &1607.8(s-) &1647.8 \\ 
 &2000 &2255.2 &2223.1 &2100.3 &2236.9(s+) &2228.6(s-) &2119.0(s-) &2181.1 \\ \hline
 TIME &500 &31.6500 &32.1350 &31.1850 &32.9450(s-) &33.1750(s-) &32.3900(s-) &47.1500 \\ 
 &1000 &124.4300 &124.1600 &117.7150 &129.8950(s-) &126.2900(s-) &120.4000(s-) &126.8800 \\ 
 &1500 &276.9850 &273.9150 &259.6000 &288.3150(s-) &276.7550(s-) &263.6800(s-) &272.3500 \\ 
 &2000 &496.1300 &494.5550 &470.5650 &515.4450(s-) &501.7350(s-) &478.2250(s-) &493.6000 \\ 
 \bottomrule
\end{tabular}
\label{tab:res_KS_fixed_gen}
\end{table} 

\begin{table}[h]
\scriptsize
\centering
\caption{Median of different measures for the feature selection experiment with fixed number of iterations.
}
\begin{tabular}{r|r|rrr|rrrr}
  \toprule
        &  \multicolumn{1}{|c}{} & \multicolumn{3}{|c}{LS} & \multicolumn{4}{|c}{LSwLL2}  \\
measure & dataset & SRP, $\alpha=2$ & SRP, $\alpha=50$ & ADP & SRP, $\alpha=2$ & SRP, $\alpha=50$ & ADP & VIGwbP\\  
\midrule
FIT &housing &0.9866 &0.9866 &0.9866 &\textbf{0.9866}(=) &\textbf{0.9866}(=) &\textbf{0.9866}(=) &\textbf{0.9866} \\ 
 &friedman &0.9873 &0.9824 &0.9849 &\textbf{0.9870}(-) &0.9821(-) &0.9842(-) &0.9866 \\ 
 &ionosphere &0.9687 &0.9687 &0.9693 &0.9687(=) &0.9687(=) &\textbf{0.9693}(=) &0.9687 \\ 
 &sonar &0.9967 &0.9950 &0.9960 &\textbf{0.9963}(-) &0.9950(=) &0.9957(-) &\textbf{0.9963} \\ 
  &cnae9m &0.9282 &0.9282 &0.9282 &\textbf{0.9282}(=) &\textbf{0.9282}(=) &\textbf{0.9282}(=) &\textbf{0.9282} \\ 
  &libras &0.8962 &0.8697 &0.8870 &0.8964(+) &0.8699(+) &0.8874(+) &\textbf{0.8967} \\ 
 &covidxr &0.8217 &0.7857 &0.8008 &0.8164(-) &0.7798(-) &0.8000(-) &\textbf{0.8221} \\ 
 &arrhythmia &0.7960 &0.7820 &0.7968 &0.7962(+) &0.7816(-) &0.8031(+) &\textbf{0.8035} \\ \hline
PELO &housing &0.4725 &0.7117 &0.7077 &0.4915(s+) &0.7257(s+) &0.7227(s+) &0.5506 \\ 
 &friedman &0.8849 &1.0000 &1.0000 &0.8448(s-) &1.0000(=) &1.0000(=) &0.9580 \\ 
 &ionosphere &0.7267 &1.0000 &0.9680 &0.6967(s-) &1.0000(=) &0.9690(+) &0.7427 \\ 
 &sonar &0.8448 &1.0000 &0.9990 &0.8258(-) &1.0000(=) &0.9990(=) &0.9219 \\ 
 &cnae9m &0.4545 &1.0000 &0.8689 &0.4434(-) &1.0000(=) &0.8729(+) &0.7588 \\  
 &libras &0.9269 &1.0000 &0.9990 &0.9209(-) &1.0000(=) &0.9990(=) &0.9780 \\ 
 &covidxr &0.8639 &1.0000 &0.9990 &0.8228(s-) &1.0000(=) &0.9990(=) &0.9630 \\ 
 &arrhythmia &0.6226 &1.0000 &0.9920 &0.5866(-) &1.0000(=) &0.9900(-) &0.9069 \\ \hline
 HDLO &housing &2.1069 &2.6525 &2.6008 &2.1165(-) &2.6865(s-) &2.6494(-) &2.2027 \\ 
 &friedman &5.0101 &45.3714 &27.9188 &4.8978(+) &45.6967(-) &28.3126(-) &8.2884 \\ 
 &ionosphere &4.0125 &11.7094 &6.8755 &3.8310(+) &11.8297(-) &6.7240(+) &4.4667 \\ 
 &sonar &4.8679 &23.5075 &14.9218 &4.7886(+) &23.8048(-) &14.7149(+) &6.8559 \\ 
 &cnae9m &3.1968 &18.2505 &5.9562 &3.2919(-) &18.5251(-) &5.9808(-) &6.9838 \\  
 &libras &5.5796 &36.3504 &23.5852 &5.4902(+) &36.6747(-) &23.0992(+) &9.0693 \\ 
 &covidxr &6.6221 &38.2112 &26.3066 &6.1987(s+) &39.1221(-) &25.9649(+) &10.8038 \\ 
 &arrhythmia &6.1887 &31.8669 &17.7629 &6.0997(+) &31.3874(+) &17.4774(+) &17.4090 \\ \hline
 HDP &housing &2.0000 &6.0000 &5.6897 &2.0000(=) &6.0000(=) &5.6507(+) &2.3654 \\ 
 &friedman &2.0000 &50.0000 &26.8468 &2.0000(=) &50.0000(=) &27.2923(-) &4.7377 \\ 
 &ionosphere &2.0000 &17.0000 &6.4985 &2.0000(=) &17.0000(=) &6.3413(+) &2.4244 \\ 
 &sonar &2.0000 &30.0000 &14.3313 &2.0000(=) &30.0000(=) &14.0641(+) &3.5826 \\ 
 &cnae9m &2.0000 &37.0000 &7.6807 &2.0000(=) &37.0000(=) &7.8959(-) &5.8959 \\  
 &libras &2.0000 &45.0000 &22.6346 &2.0000(=) &45.0000(=) &22.1852(+) &4.8809 \\ 
 &covidxr &2.0000 &46.0000 &24.9219 &2.0000(=) &46.0000(=) &24.8458(+) &5.1491 \\ 
 &arrhythmia &2.0000 &50.0000 &16.7678 &2.0000(=) &50.0000(=) &16.6867(+) &13.3794 \\ \hline
FDP &housing &0.0042 &0.0115 &0.0108 &0.0042(+) &0.0115(-) &0.0108(+) &0.0254 \\ 
 &friedman &0.0042 &0.0181 &0.0151 &0.0042(-) &0.0181(-) &0.0150(+) &0.0069 \\ 
 &ionosphere &0.0535 &0.1066 &0.0877 &0.0524(+) &0.1056(+) &0.0865(+) &0.0600 \\ 
 &sonar &0.0873 &0.1730 &0.1551 &0.0844(+) &0.1721(+) &0.1549(+) &0.1180 \\ 
 &cnae9m &0.0302 &0.2831 &0.1028 &0.0305(-) &0.2825(+) &0.1067(-) &0.1022 \\ 
 &libras &0.0728 &0.1743 &0.1593 &0.0735(-) &0.1750(-) &0.1596(-) &0.1046 \\ 
 &covidxr &0.0798 &0.1939 &0.1846 &0.0778(+) &0.1916(+) &0.1812(+) &0.1271 \\ 
 &arrhythmia &0.0410 &0.1524 &0.1205 &0.0398(+) &0.1523(+) &0.1209(-) &0.1017 \\  \hline
 FHRP &housing &0.0021 &0.0019 &0.0019 &0.0021(+) &0.0019(-) &0.0019(-) &0.0113 \\ 
 &friedman &0.0021 &0.0004 &0.0006 &0.0021(-) &0.0004(-) &0.0005(+) &0.0014 \\ 
 &ionosphere &0.0268 &0.0063 &0.0134 &0.0262(+) &0.0062(+) &0.0134(-) &0.0239 \\ 
 &sonar &0.0436 &0.0058 &0.0109 &0.0422(+) &0.0057(+) &0.0108(+) &0.0327 \\ 
 &cnae9m &0.0151 &0.0077 &0.0132 &0.0153(-) &0.0076(+) &0.0134(-) &0.0173 \\  
 &libras &0.0364 &0.0039 &0.0071 &0.0368(-) &0.0039(-) &0.0072(-) &0.0217 \\ 
 &covidxr &0.0399 &0.0042 &0.0073 &0.0389(+) &0.0042(+) &0.0072(+) &0.0246 \\ 
 &arrhythmia &0.0205 &0.0030 &0.0073 &0.0199(+) &0.0030(+) &0.0071(+) &0.0076 \\ \hline
NILS &housing &23.4 &30.2 &29.7 &25.2(s-) &33.8(s-) &32.9(s-) &27.6 \\ 
 &friedman &207.4 &341.5 &304.8 &206.9(+) &355.6(s-) &315.9(s-) &242.9 \\ 
 &ionosphere &68.8 &101.8 &83.4 &70.1(s-) &109.5(s-) &86.8(s-) &73.7 \\ 
 &sonar &121.7 &193.7 &165.2 &123.4(s-) &210.4(s-) &174.1(s-) &135.4 \\ 
 &cnae9m &148.2 &262.6 &195.3 &167.1(s-) &324.8(s-) &232.1(s-) &220.3 \\ 
 &libras &185.6 &314.8 &266.9 &191.2(s-) &338.5(s-) &279.0(s-) &220.0 \\ 
 &covidxr &222.2 &315.0 &290.0 &228.9(s-) &351.2(s-) &319.2(s-) &262.2 \\ 
 &arrhythmia &604.7 &1017.4 &839.9 &669.0(s-) &1224.2(s-) &987.5(s-) &872.4 \\ \hline
TIME &housing &114.9900 &156.0500 &152.4700 &125.8200(-) &175.4000(s-) &170.3200(s-) &128.0500 \\ 
 &friedman &4490.0500 &16544.4300 &11854.2000 &4403.6300(+) &16949.0900(-) &12916.8200(-) &6309.0200 \\ 
 &ionosphere &261.4300 &542.4700 &384.3600 &274.1600(s-) &612.1600(s-) &410.8800(s-) &301.4500 \\ 
 &sonar &304.2800 &651.4800 &565.9000 &318.9500(s-) &716.4500(s-) &574.8400(-) &346.7400 \\ 
 &cnae9m &1825.3500 &3362.8800 &2402.8200 &2072.9000(s-) &4214.0100(s-) &2835.7200(s-) &2796.4600 \\ 
 &libras &1990.8400 &4567.5000 &3917.5300 &2011.7800(-) &5065.4500(s-) &4174.8300(s-) &2608.9200 \\ 
 &covidxr &4830.4800 &8299.1100 &7030.7800 &4839.9100(-) &9219.6200(s-) &7749.9900(s-) &6029.8800 \\ 
 &arrhythmia &12854.9700 &28962.1600 &17893.5700 &13736.0800(-) &35984.9000(s-) &21484.9200(s-) &20050.5600 \\
 \bottomrule
\end{tabular}
\label{tab:res_FS_fixed_gen}
\end{table} 

\begin{table}[h]
\scriptsize
\centering
\caption{Median of different measures for the NK landscapes experiment with fixed time.
}
\begin{tabular}{r|rrr|rrr|rrrr}
  \toprule
        &  \multicolumn{3}{|c}{} & \multicolumn{3}{|c}{LS} & \multicolumn{4}{|c}{LSwLL2}  \\
measure & model & $N$ & $k$ & SRP, $\alpha=2$ & SRP, $\alpha=50$ & ADP & SRP, $\alpha=2$ & SRP, $\alpha=50$ & ADP & VIGwbP\\  
\midrule
FIT & adjacent &100 &3 &0.7523 &0.7497 &0.7523 &\textbf{0.7523}(=) &0.7496(-) &\textbf{0.7523}(=) &\textbf{0.7523} \\ 
 &  &  &5 &\textbf{0.7765} &0.7547 &0.7692 &\textbf{0.7765}(=) &0.7536(-) &0.7684(-) &\textbf{0.7765} \\ 
 &  &500 &3 &0.7450 &0.7422 &0.7461 &0.7449(-) &0.7420(-) &0.7460(-) &\textbf{0.7466} \\ 
 &  &  &5 &0.7710 &0.7542 &0.7714 &0.7709(-) &0.7549(+) &0.7710(-) &\textbf{0.7744} \\ 
 &  &1000 &3 &0.7457 &0.7413 &0.7461 &0.7456(-) &0.7406(s-) &0.7463(+) &\textbf{0.7468} \\ 
 &  &  &5 &0.7718 &0.7547 &0.7711 &0.7717(-) &0.7546(-) &0.7700(-) &\textbf{0.7768} \\ 
 & random &100 &3 &0.7538 &0.7518 &0.7538 &\textbf{0.7538}(=) &0.7518(=) &\textbf{0.7538}(=) &\textbf{0.7538} \\ 
 &  &  &5 &0.7746 &0.7815 &0.7861 &0.7730(-) &0.7808(-) &0.7861(+) &\textbf{0.7862} \\ 
 &  &500 &3 &0.7489 &0.7508 &0.7506 &0.7485(-) &0.7503(-) &0.7505(-) &\textbf{0.7507} \\ 
 &  &  &5 &0.7675 &0.7775 &0.7723 &0.7681(+) &\textbf{0.7768}(-) &0.7710(-) &0.7732 \\ 
 &  &1000 &3 &0.7465 &0.7470 &0.7475 &0.7466(+) &0.7464(s-) &0.7474(-) &\textbf{0.7476} \\ 
 &  &  &5 &0.7685 &0.7769 &0.7708 &0.7677(-) &\textbf{0.7742}(-) &0.7708(-) &0.7719 \\ \hline
 NI & adjacent &100 &3 &85269.5 &49690.0 &68812.0 &76746.5(s-) &39379.5(s-) &56456.5(s-) &70244.5 \\ 
 &  &  &5 &89829.0 &51788.5 &64293.0 &82756.5(s-) &41867.5(s-) &55082.5(s-) &70466.0 \\ 
 &  &500 &3 &17224.0 &10380.5 &13646.0 &15530.5(s-) &8253.5(s-) &11477.5(s-) &14515.0 \\ 
 &  &  &5 &16403.0 &9207.5 &12779.0 &15179.5(s-) &7423.0(s-) &10502.0(s-) &13557.5 \\ 
 &  &1000 &3 &8001.5 &5057.0 &6380.0 &7363.5(s-) &4019.5(s-) &5292.0(s-) &6978.0 \\ 
 &  &  &5 &8339.0 &4672.0 &6207.5 &7655.5(s-) &3675.5(s-) &5087.0(s-) &6756.5 \\ 
 & random &100 &3 &84266.0 &47102.0 &61802.0 &73843.5(s-) &37690.0(s-) &49731.0(s-) &56692.0 \\ 
 &  &  &5 &85266.5 &43138.0 &48996.0 &77965.0(s-) &34943.0(s-) &41562.0(s-) &41380.0 \\ 
 &  &500 &3 &16966.0 &10001.5 &12486.0 &14983.5(s-) &7816.5(s-) &10198.5(s-) &11678.5 \\ 
 &  &  &5 &16663.0 &7871.0 &11388.5 &14819.5(s-) &6249.0(s-) &9169.5(s-) &8153.5 \\ 
 &  &1000 &3 &7992.0 &4971.5 &5797.5 &6942.0(s-) &3776.0(s-) &4712.5(s-) &5813.0 \\ 
 &  &  &5 &7960.5 &4234.0 &5672.5 &6917.0(s-) &3243.5(s-) &4504.0(s-) &4272.0 \\ 
 \bottomrule
\end{tabular}
\label{tab:res_NK_fixed_time}
\end{table} 

\begin{table}[h]
\scriptsize
\centering
\caption{Median of different measures for the 0-1 knapsack problem experiment with fixed time.
}
\begin{tabular}{r|r|rrr|rrrr}
  \toprule
        &  \multicolumn{1}{|c}{} & \multicolumn{3}{|c}{LS} & \multicolumn{4}{|c}{LSwLL2}  \\
measure & $N$ & SRP, $\alpha=2$ & SRP, $\alpha=50$ & ADP & SRP, $\alpha=2$ & SRP, $\alpha=50$ & ADP & VIGwbP\\  
\midrule
ERR &500 &0.0006 &0.0902 &0.1013 &\textbf{0.0006}(-) &0.0912(-) &0.0974(+) &0.0046 \\ 
 &1000 &0.0013 &0.0969 &0.1152 &\textbf{0.0015}(s-) &0.0981(-) &0.1170(-) &0.0163 \\ 
 &1500 &0.0025 &0.1027 &0.1240 &\textbf{0.0028}(s-) &0.1041(-) &0.1230(+) &0.0435 \\ 
 &2000 &0.0039 &0.1054 &0.1304 &\textbf{0.0043}(s-) &0.1076(s-) &0.1316(-) &0.0617 \\  \hline
 NI &500 &224304.0 &224175.0 &231960.5 &193585.0(s-) &191461.0(s-) &200688.5(s-) &158861.5 \\ 
 &1000 &116056.0 &116444.0 &122201.5 &98227.0(s-) &102561.0(s-) &106490.0(s-) &91704.0 \\ 
 &1500 &78595.5 &78826.0 &84259.0 &65521.5(s-) &70143.5(s-) &74143.0(s-) &71011.5 \\ 
 &2000 &58868.5 &59511.5 &63821.5 &50551.5(s-) &53551.5(s-) &56256.0(s-) &55774.5 \\ 
 \bottomrule
\end{tabular}
\label{tab:res_KS_fixed_time}
\end{table} 
 
\begin{table}[h]
\scriptsize
\centering
\caption{Median of different measures for the feature selection experiment with fixed time.
}
\begin{tabular}{r|r|rrr|rrrr}
  \toprule
        &  \multicolumn{1}{|c}{} & \multicolumn{3}{|c}{LS} & \multicolumn{4}{|c}{LSwLL2}  \\
measure & dataset & SRP, $\alpha=2$ & SRP, $\alpha=50$ & ADP & SRP, $\alpha=2$ & SRP, $\alpha=50$ & ADP & VIGwbP\\  
\midrule
FIT &housing &0.9866 &0.9866 &0.9866 &\textbf{0.9866}(=) &\textbf{0.9866}(=) &\textbf{0.9866}(=) &\textbf{0.9866} \\ 
 &friedman &0.9869 &0.9814 &0.9848 &\textbf{0.9870}(+) &0.9811(-) &0.9842(-) &0.9863 \\ 
 &ionosphere &0.9693 &0.9693 &0.9693 &\textbf{0.9693}(=) &\textbf{0.9693}(=) &\textbf{0.9693}(=) &\textbf{0.9693} \\ 
 &sonar &0.9967 &0.9953 &0.9960 &0.9963(s-) &0.9950(-) &0.9960(=) &\textbf{0.9967} \\ 
  &cnae9m &0.9282 &0.9282 &0.9282 &\textbf{0.9282}(=) &\textbf{0.9282}(=) &\textbf{0.9282}(=) &\textbf{0.9282} \\ 
  &libras &0.8962 &0.8679 &0.8868 &0.8964(+) &0.8677(-) &0.8870(+) &\textbf{0.8967} \\ 
 &covidxr &0.8154 &0.7766 &0.7980 &0.8095(-) &0.7720(-) &0.7941(-) &\textbf{0.8156} \\ 
 &arrhythmia &0.7960 &0.7678 &0.7895 &\textbf{0.7962}(+) &0.7741(+) &\textbf{0.7962}(+) &0.7895 \\ \hline
NI &housing &4131.0 &3060.0 &3306.0 &3968.0(-) &2831.0(s-) &2806.0(s-) &3502.0 \\ 
 &friedman &747.0 &215.0 &338.0 &780.0(+) &199.0(s-) &313.0(-) &542.0 \\ 
 &ionosphere &3430.0 &1437.0 &2203.0 &3169.0(s-) &1219.0(s-) &2208.0(+) &3196.0 \\ 
 &sonar &3240.0 &1254.0 &1702.0 &2889.0(s-) &1059.0(s-) &1514.0(s-) &2570.0 \\ 
 &cnae9m &1565.0 &734.0 &1142.0 &1386.0(s-) &590.0(s-) &967.0(s-) &1025.0 \\ 
 &libras &1177.0 &443.0 &611.0 &1142.0(-) &393.0(s-) &575.0(-) &876.0 \\  
 &covidxr &688.0 &335.0 &433.0 &665.0(-) &296.0(s-) &388.0(s-) &550.0 \\ 
 &arrhythmia &772.0 &303.0 &500.0 &695.0(-) &246.0(s-) &407.0(s-) &407.0 \\ 
 \bottomrule
\end{tabular}
\label{tab:res_FS_fixed_time}
\end{table}

\subsection{Results: weighted variable interaction graphs}
 \label{sec:res_vig}
 An advantage of LSwLL2 over LS is that it builds an empirical weighted variable interaction graph as a side effect of local search. 
 We show here some examples of using the VIGw for helping generating insights about the problems and optimizers. 
 
  First, we analyze the number of edges of the empirical VIGw discovered by ILS with LSwLL2. 
One can remember that LSwLL2 never returns a false linkage (Corollary~\ref{corollary:LSwLL}). 
In other words, if two vertices are connected in the empirical VIGw, they are also connected in the VIGw. 
In the NK landscapes, the VIG is known \emph{a priori}. 
Thus, Table~\ref{tab:nedges_NK_fixed_gen} shows the percentage of the edges of the VIGw in the empirical VIGw for the experiment with fixed number of iterations. 
The results show that, after 5000 iterations, ILS with LSwLL2 was able to find at least $90.4\%$ of the edges of the VIGw. 
Regarding the algorithms, the worse results are for SRP with $\alpha=2$, i.e., when the perturbation strength is the weakest and the number of iterations of LS ($NILS$) is the smallest (Table~\ref{tab:res_NK_fixed_gen_2}). 
Regarding the instances, the worse results are for $N=1000$ and $k=3$. 
More variables (higher $N$) mean that the VIGw has more edges. 
The VIGw has more edges for $k=5$ than for $k=3$; however, for $k=3$, a bit flip will impact a fewer number of subfunctions in Eq.~\eqref{eq:eval_NK} (see also $FDP$ in Table~\ref{tab:res_NK_fixed_gen_1}). 
This explains the worse edge-discovery results for $k=3$ in NK landscapes. 

Tables \ref{tab:nedges_KS_fixed_gen} and \ref{tab:nedges_FS_fixed_gen} show the percentage of the edges of the empirical VIGw over all possible edges for the other two problems in the experiments with fixed number of iterations. 
Specially for the feature selection problem, the graph is dense, i.e., most of the possible edges are present in the VIGw. 
In these cases, knowing the VIG is of little help. 
However, the weights of the VIGw can be useful when designing operators and analyzing the optimization instances and algorithms. 

\subsubsection{NK Landscapes: }
\begin{figure}[h]
  \centering
  \includegraphics[width=0.49\linewidth]{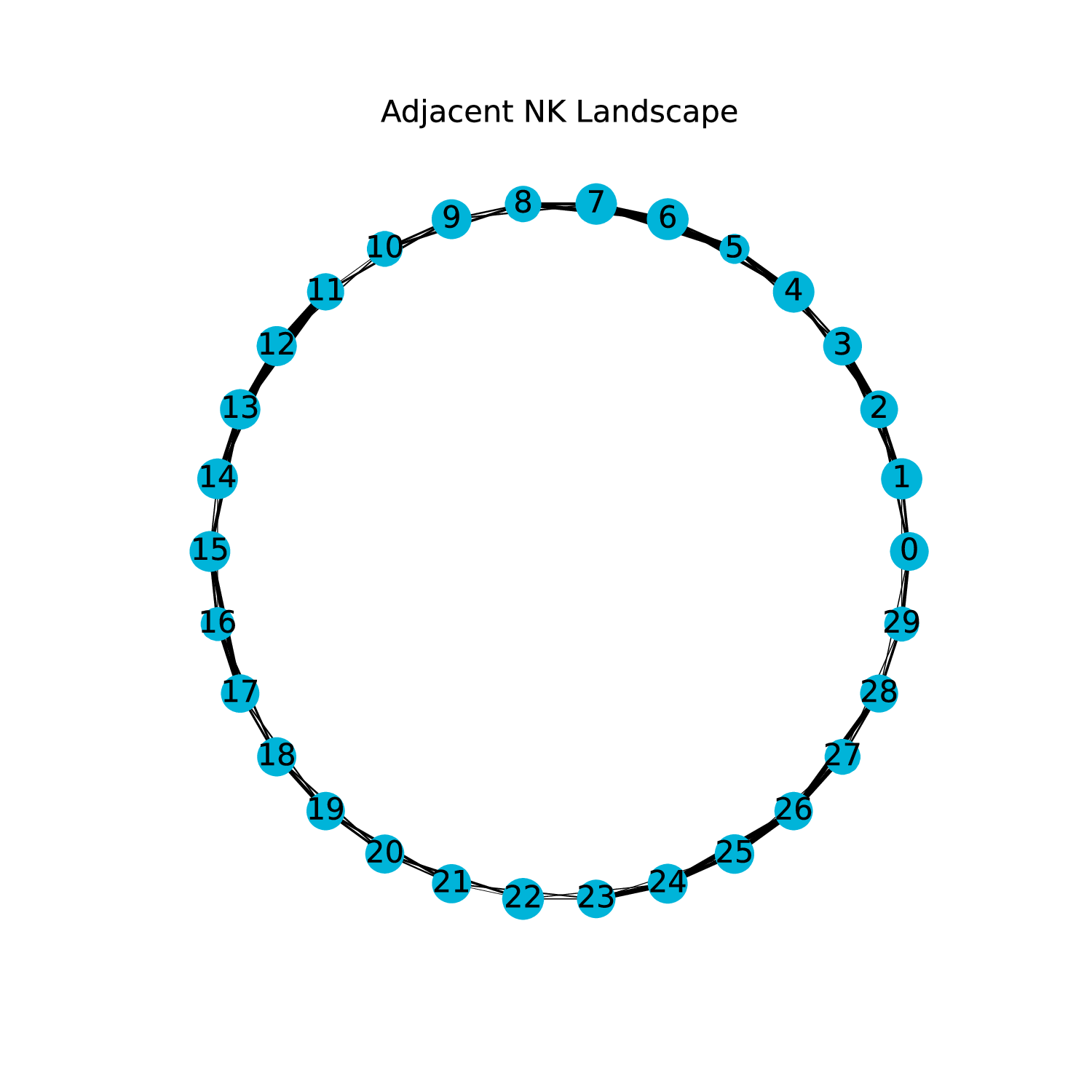}  
   \includegraphics[width=0.49\linewidth]{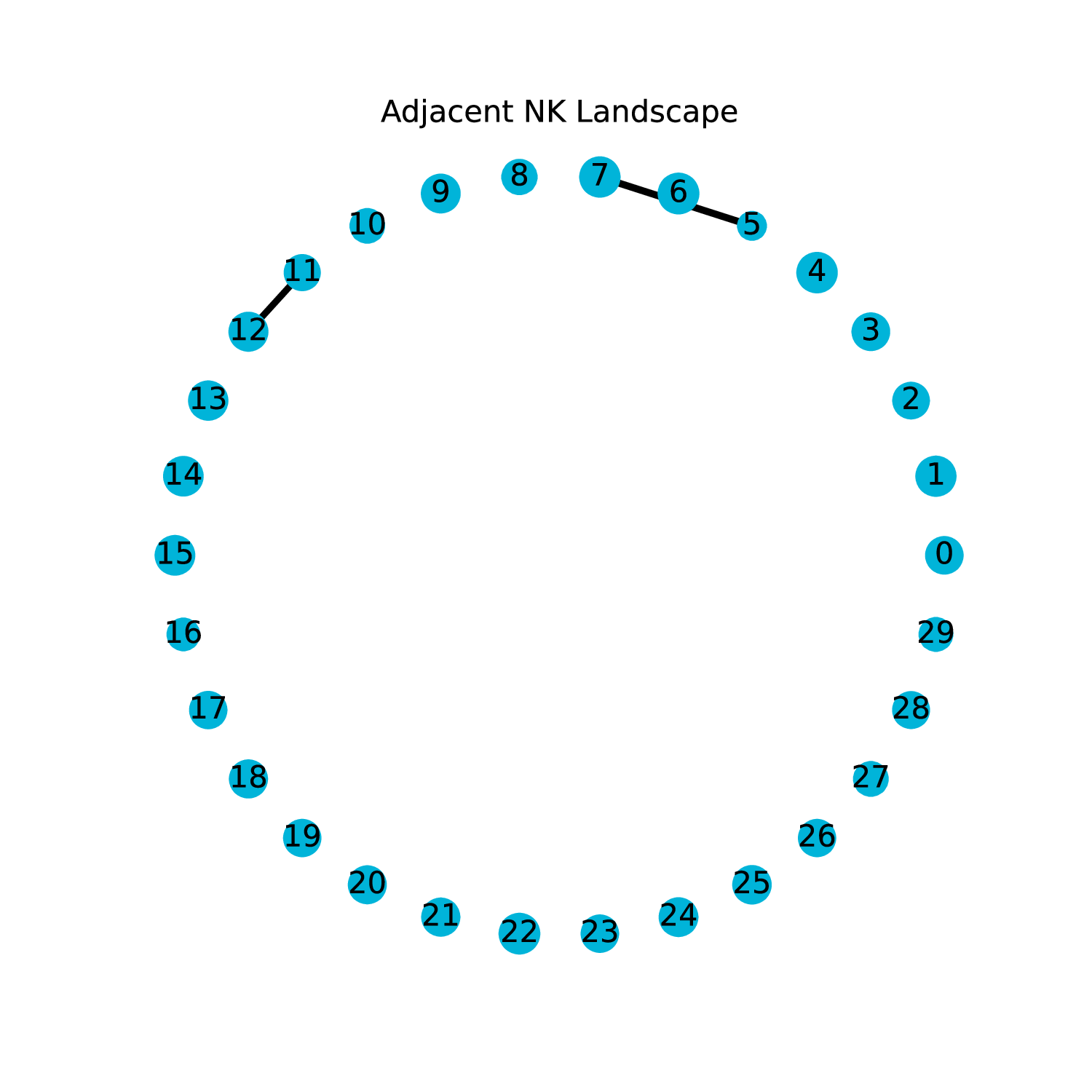} \\
     \includegraphics[width=0.49\linewidth]{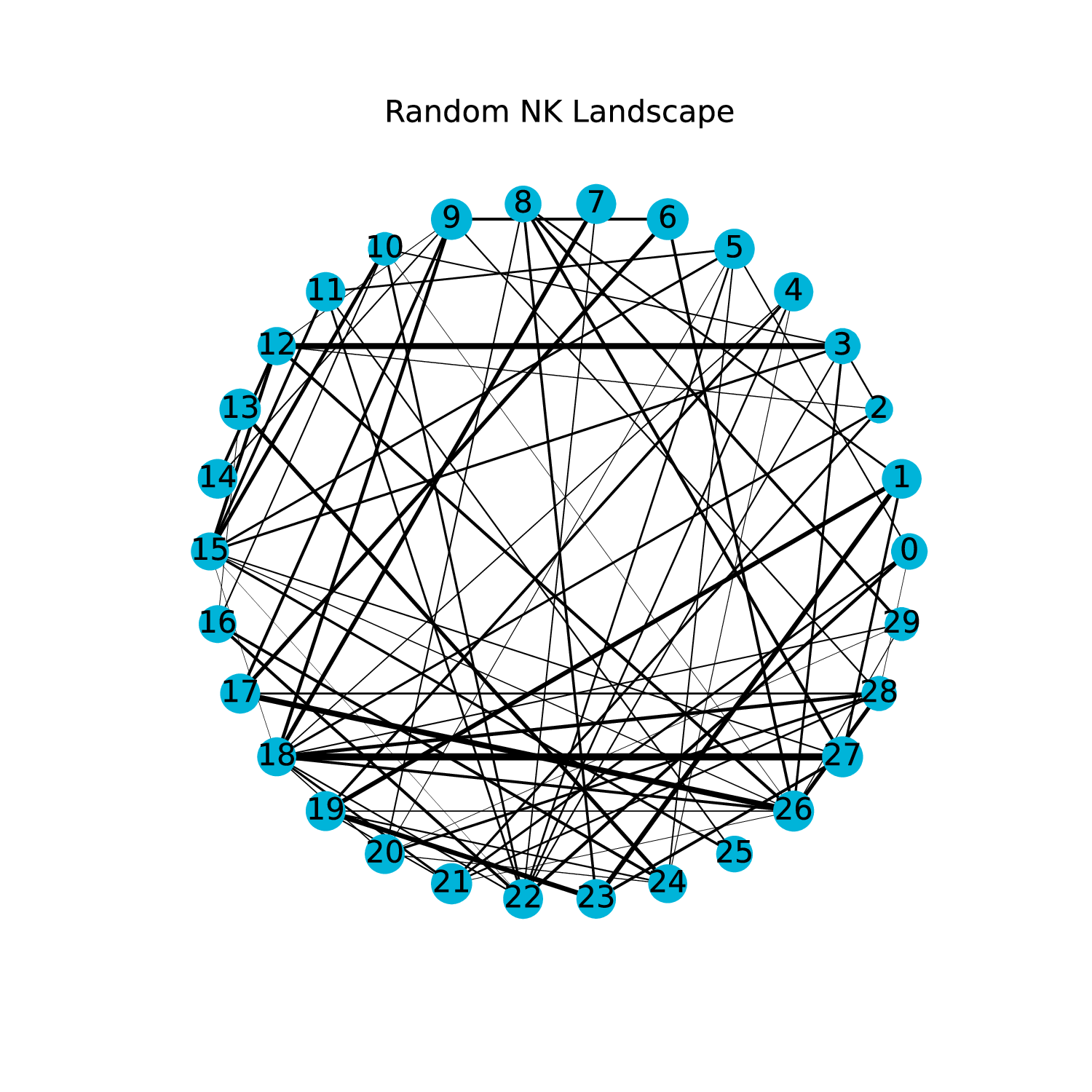}
   \includegraphics[width=0.49\linewidth]{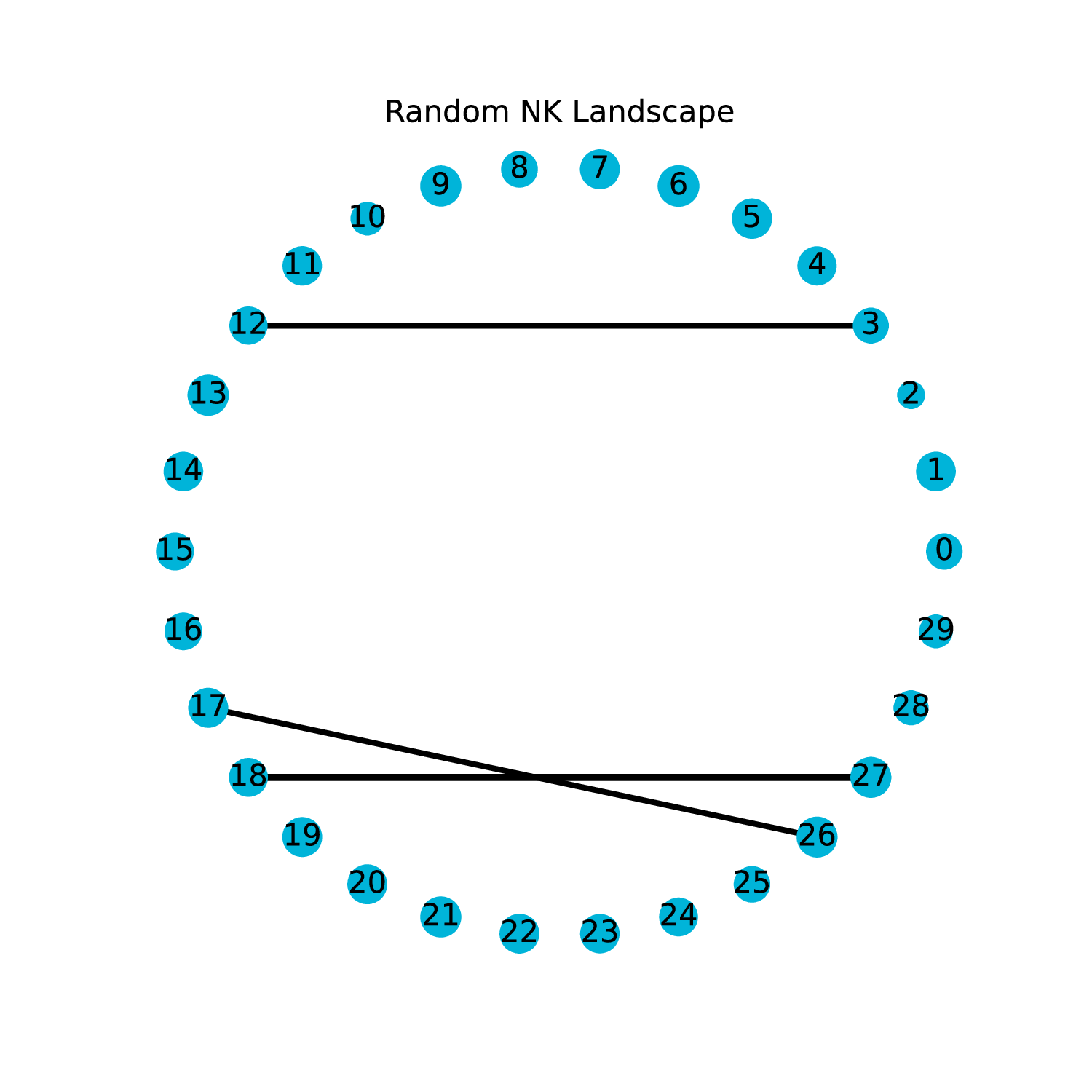}
  \caption{Empirical weighted VIG found by ILS with VIGwbP and LSwLL2 in one run for the NK landscapes ($N=30$, $k=3$) experiment with fixed number of iterations. The widths of the lines are proportional to the weights of the respective edges. The size of the i-th node is proportional to the contribution  $f_i$ for the evaluation of the best individual found by ILS. Left: complete graph. Right: only the edges with largest weights are presented. The largest weights are defined according to the procedure to compute the threshold $\beta$ in Section~\ref{sec:VIGwbP}.}
  \label{fig:VIG_NK}
\end{figure}

Figure~\ref{fig:VIG_NK} shows the empirical VIGw found by ILS with VIGwbP in experiments with one run of the adjacent and random models for $N=30$ and $k=3$. 
The empirical VIGw of only one instance for each model is presented. 
We added additional experiments with $N=30$ because it is easier to analyze the weighted graphs for small dimensions of the problem, i.e., for a small number of graph vertices. 
The VIG indicates when two variables interact, but does not show information about the strength of the interactions. 
From Figure~\ref{fig:VIG_NK}, we can observe the strength of the interactions, inferred by LSwLL2, in the empirical VIGw. 
When flipping two variables generally cause small variation in the evaluation of the subfunctions in Eq~\eqref{eq:eval_NK}, then the weight between the respective vertices in the VIGw is small. 
Otherwise, the weight is large. 
Thus, the impact of flipping two variables depends on the number of subfunctions that these variables appear together in Eq~\eqref{eq:eval_NK} and also in the difference between the minimum and maximum values that the evaluation of these subfunctions can assume.  
One can remember that, in NK landscapes, the evaluation of the subfunctions for each combination of variables is randomly generated for each instance. 
The edges with largest weights in examples with adjacent and random NK landscapes can be seen in Figure~\ref{fig:VIG_NK}. 
For the random model, we can observe that variables with the largest weights, e.g., $x_{26}$, impact subfunctions with largest contribution to the evaluation of the best individual. 

\subsubsection{Knapsack Problem: }
\begin{figure}[h]
  \centering
  \includegraphics[width=0.49\linewidth]{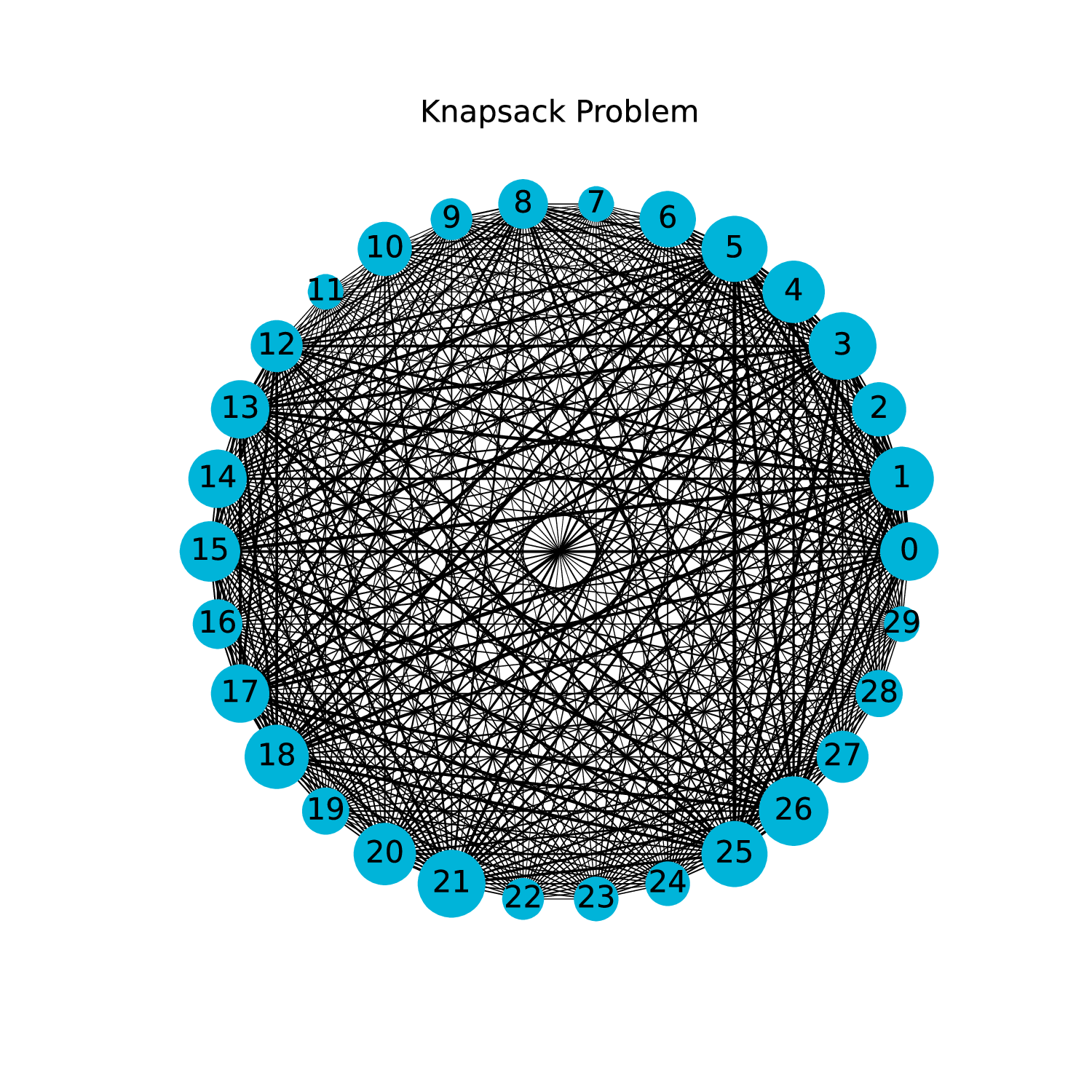}
   \includegraphics[width=0.49\linewidth]{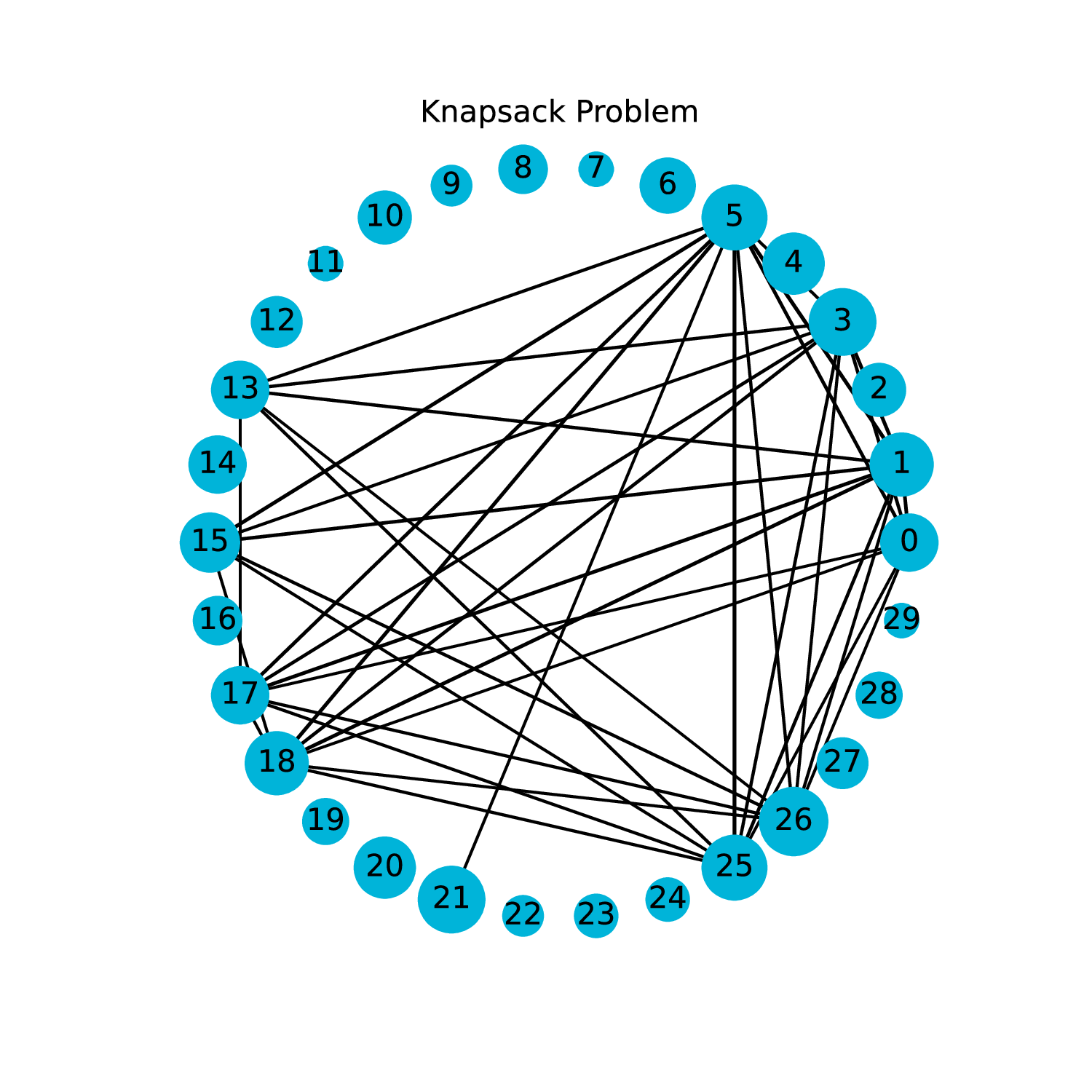} 
  \caption{Empirical weighted VIG found by ILS with VIGwbP and LSwLL2 in one run for the 0-1 knapsack problem ($N=30$) experiment with fixed number of iterations. Here, the size of the i-th node is proportional to the weight of the i-th object of the knapsack problem (heavier objects are represented by larger circles). Left: complete graph. Right: only the edges with largest weights are presented.  }
  \label{fig:VIG_KS}
\end{figure}

Figure~\ref{fig:VIG_KS} helps to understand how the interaction of variables, inferred by the weighs, impacts the search. 
In this example, the empirical VIGw found in a run of ILS with VIGwbP in a random instance of the 0-1 knapsack problem with $N=30$ is presented. 
The VIG is dense in this example, and its analysis provides little information. 
However, analyzing the largest weights of the empirical VIGw can help understanding the optimization process. 
The edges with largest weights in Figure~\ref{fig:VIG_KS} are generally those connected to vertices that represent heavy objects in the knapsack problem. 
This is a result of the penalty function (Eq.~\ref{eq:knap_p}) used in the knapsack problem. 
The largest variations in the evaluation of a solution (Eq.~\ref{eq:knap_f}) are generally caused by adding or removing heavy objects because this will often impact the penalty applied to the solution. 

\subsubsection{Feature Selection Problem: }
ILS with LSwLL2 produces a relevant information during feature selection:  \emph{feature interaction}, or \emph{variable interaction}. 
\citet{inglis2022} define variable interaction as a measure (scalar quantity) that expresses the degree to which two or more variables combine to affect the dependent variable.  
Here, feature interaction between variables $x_h$ and $x_g$ is indicated by the weight of edge ($h$,$g$) of the empirical VIGw. 
Thus, given a dataset and machine learning model, the empirical VIGw can be employed for the visualization of all two-variable interactions. 
\citet{inglis2022} proposed a variable interaction visualization in the form of heatmaps and graphs (network plot). 
In a network plot, nodes represent variables and weighted edges represent the interaction between the variables. 
The weights of the network plot are computed based on Friedman's \emph{H-statistic} or \emph{H-index}~\cite{friedman2008}. 
\citet{friedman2008} say that there is interaction between two variables, $x_h$ and $x_g$, if the impact on the output function $f(\mathbf{x})$ as a result of changing $x_h$ depends on $x_g$. 
Given a machine learning model and a dataset, the H-statistic for each pair of variables measures the change in the predicted value of the model as variables vary over their marginal distribution. 
Detecting variable interaction and computing the H-statistic in ~\cite{friedman2008,inglis2022} is similar to the strategies proposed here for the LSwLL2 (Section~\ref{sec:LSwLL}). 
However, the H-statistic needs many evaluations  of a machine learning model to compute the variable interaction for each pair of variables. 
Feature selection is not done while finding the interactions between variables. 
On the other hand, LSwLL2 needs no additional solution evaluations while selecting the features. 

\begin{figure}[h]
  \centering
  \includegraphics[width=0.49\linewidth]{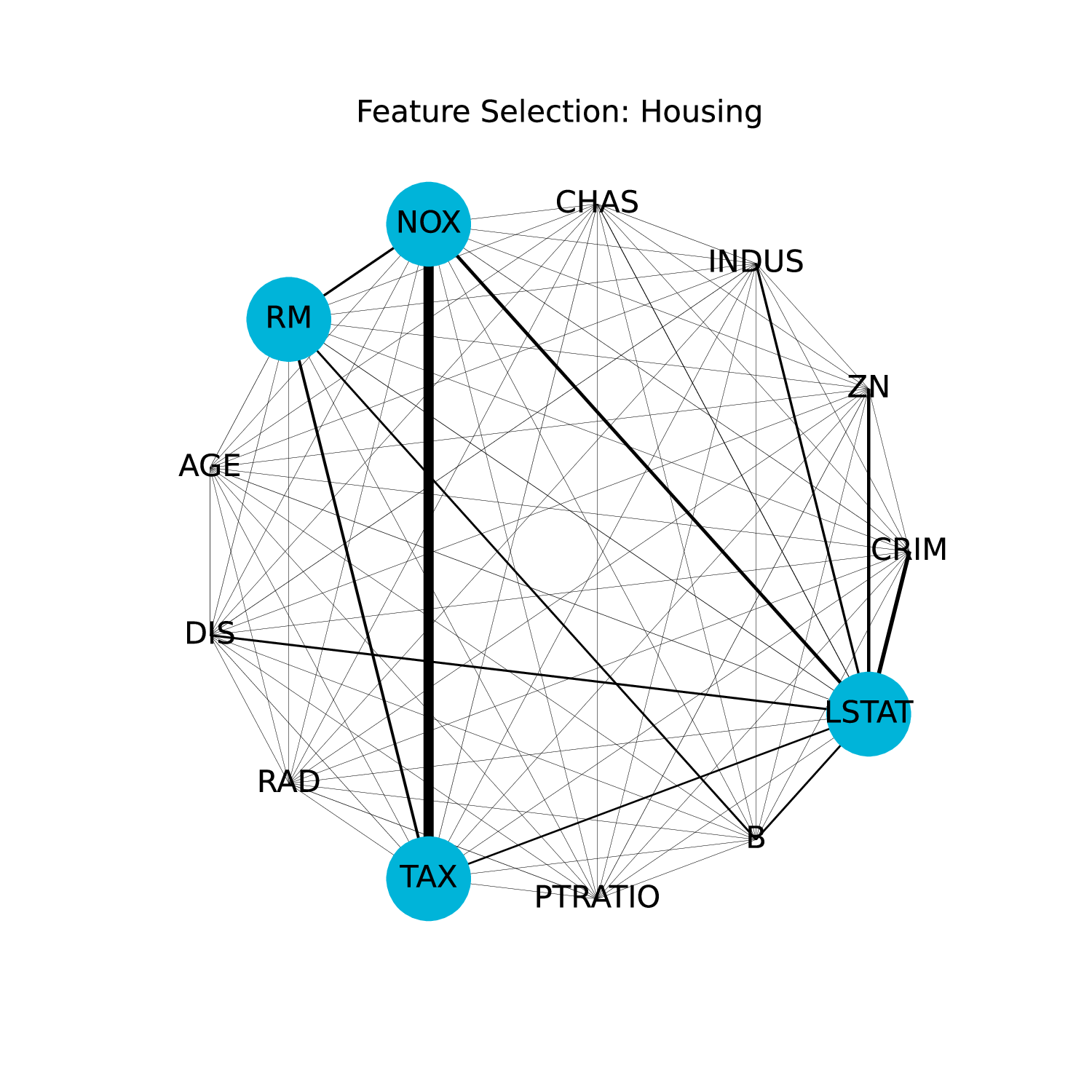}
   \includegraphics[width=0.49\linewidth]{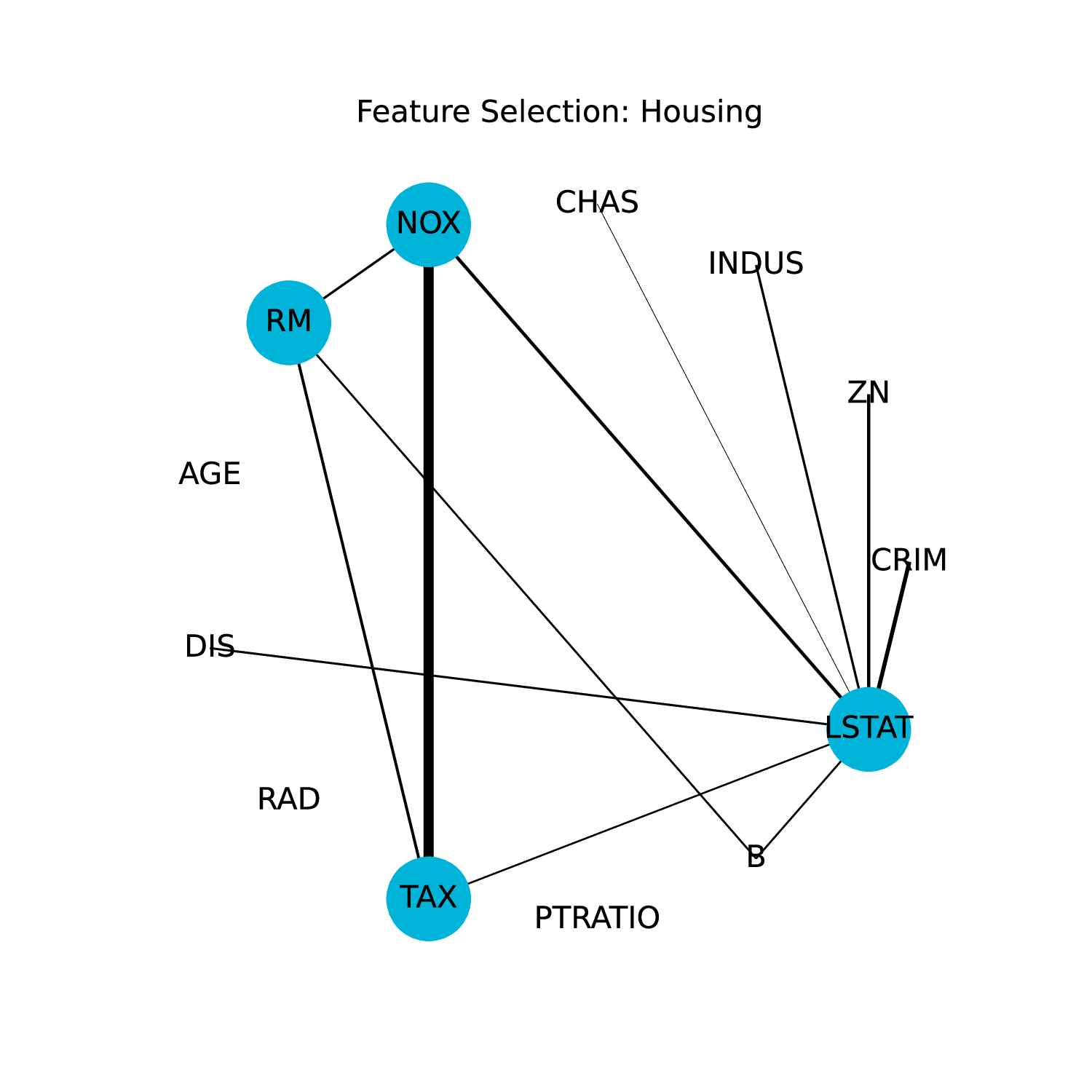} 
  \caption{Empirical weighted VIG found by ILS with VIGwbP and LSwLL2 in the first run of the experiments with fixed number of iterations for the feature selection problem with dataset housing. Here, the features selected by ILS are indicated by the blue circles. Left: complete graph. Right: only the edges with largest weights are presented.}
  \label{fig:VIG_FS_housing}
\end{figure}

\begin{figure}[h]
  \centering
  \includegraphics[width=0.49\linewidth]{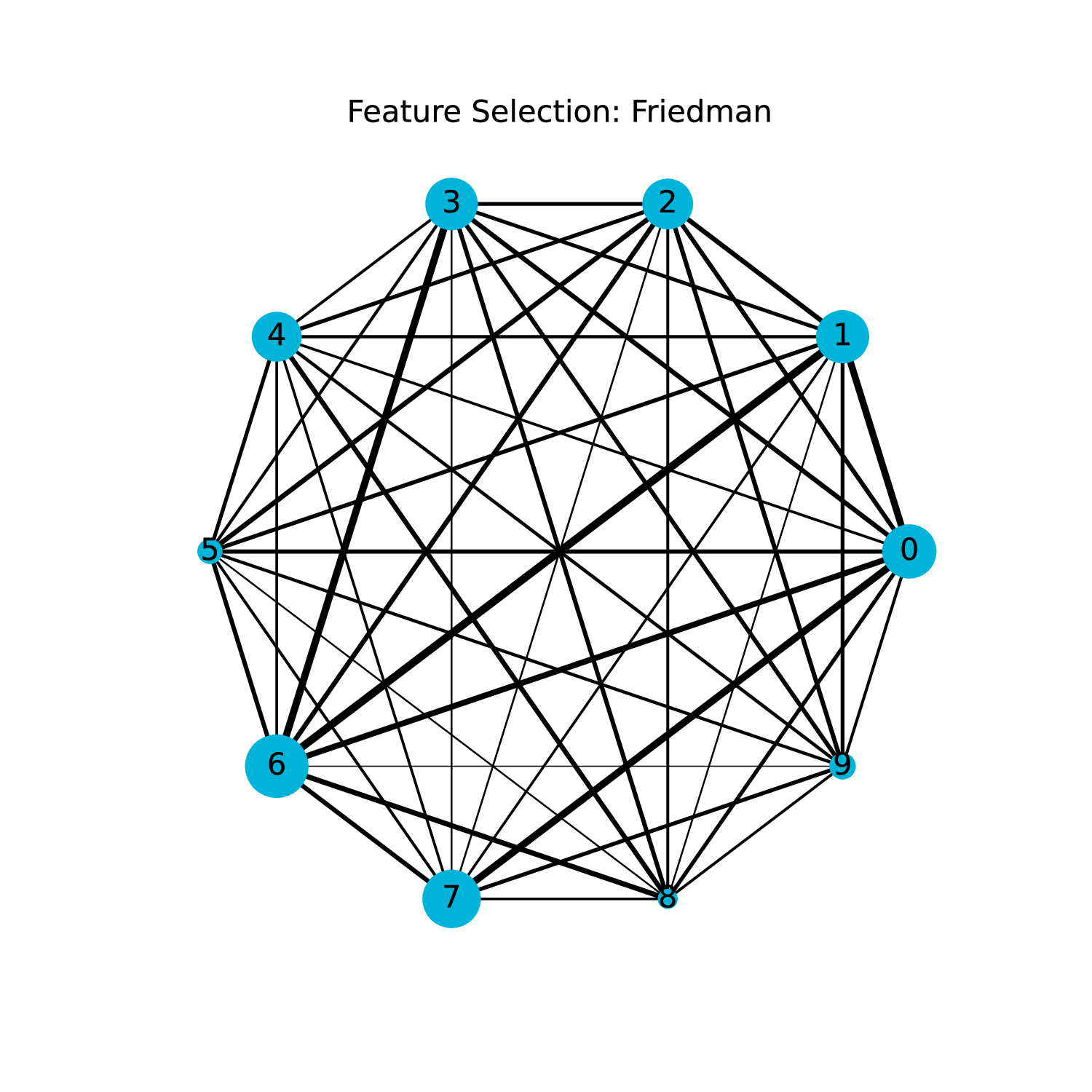}
   \includegraphics[width=0.49\linewidth]{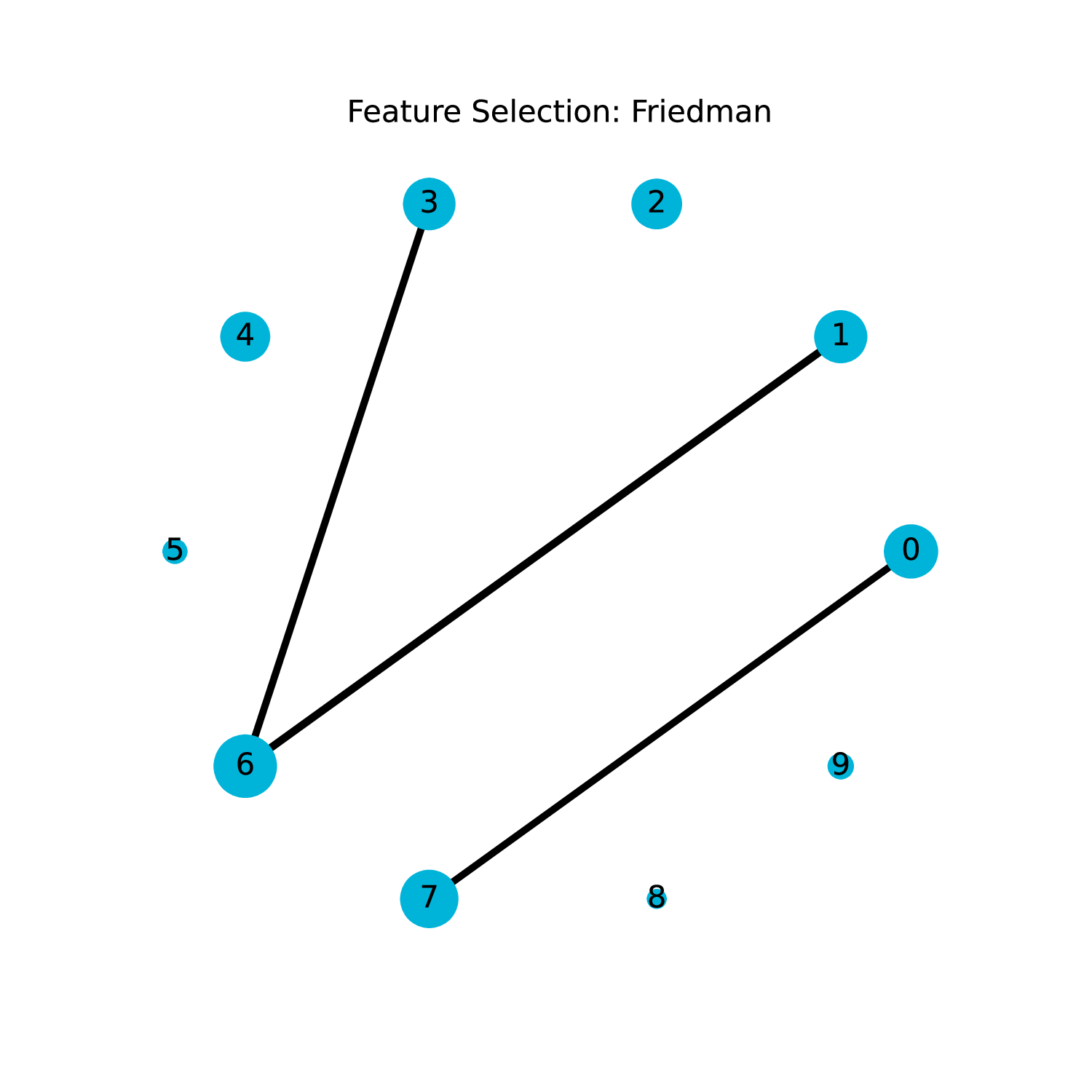} 
  \caption{Subgraph (with the 10 first nodes) of the VIGw found by ILS with VIGwbP and LSwLL2 in the first run of the experiments with fixed number of iterations for the feature selection problem with dataset friedman. Here, the size of the i-th node is proportional to the correlation between the input $u_i$ and the desired output $F(\mathbf{u})$ (Eq.~\ref{eq:f_friedman}) for the dataset. Features $x_5$, $x_8$, and $x_9$ were not selected by ILS. Left: complete graph. Right: only the edges with largest weights are presented.}
  \label{fig:VIG_FS_friedman}
\end{figure}

Figures \ref{fig:VIG_FS_housing} and \ref{fig:VIG_FS_friedman} respectively show the VIGws obtained in the first run of ILS with VIGwbP and LSwLL2 in the experiment with fixed number of iterations for datasets housing and friedman. 
In Figure~\ref{fig:VIG_FS_friedman}, only the subgraph with the 10 first decision variables is presented. 
In the housing dataset, the goal is to predict the median house value in $n_{ex}=506$ neighborhoods of the Boston metropolitan area. 
In Figure \ref{fig:VIG_FS_housing}, the features selected by the best solution of ILS were: NOX (nitric oxides concentration), RM (average number of rooms per dwelling), LSTAT (percentage of lower status population), and TAX (property-tax rate).  
In~\cite{friedman2008}, using a different machine learning model (RuleFIT), the three features with highest relative importance were NOX, RM, and LSTAT. 
TAX is the sixth variable with highest relative importance. 
Regarding variable interaction, Figure~\ref{fig:VIG_FS_housing} shows that the most important interactions involve features selected by the best solution of ILS. 
The strongest interaction is between NOX and TAX. 
LSTAT is the variable with more strong interactions; a similar result was obtained by~\citet{friedman2008}. 

For the experiment with friedman dataset, features $x_5$, $x_8$, and $x_9$ were not selected by ILS, indicating that they are not important for the machine learning model.  
In Eq.~\eqref{eq:f_friedman}, $F(\mathbf{u})$ does not depend on $u_8$, and $u_9$. 
However, it depends on inputs $u_0$ to $u_7$. 
The explanation for not selecting $x_5$ is given by the impact of $u_5$ on the output $F(\mathbf{u})$ for the dataset with $n_{ex}=500$ and the machine learning model (KNN). 
This can be seen in Figure~\ref{fig:VIG_FS_friedman}, where the sizes of the nodes are proportional to the correlation between their respective inputs and the output $F(\mathbf{u})$. 
The correlation for variable $u_5$ is the weakest among inputs $u_0$ to $u_7$. 
In fact, the strongest interactions in the empirical VIGw are between variables with strongest correlations to $F(\mathbf{u})$. 
However, it is important to observe that the results are dependent on the size of the dataset and the machine learning model. 
Here, the number of examples in the dataset is $n_{ex}=500$, while $5000$ examples were generated by~\citet{friedman2008}. 
In the results of the experiment with KNN and \emph{random forests} presented by \citet{inglis2022}, more variable interactions appeared when KNN was used.

\begin{table}[h]
\scriptsize
\centering
\caption{Median of the percentage of the edges of the VIGw in the empirical VIGw for the NK landscapes experiment with fixed number of iterations.}
\begin{tabular}{rrr|rrrr}
  \toprule
model & $N$ & $k$ & SRP, $\alpha=2$ & SRP, $\alpha=50$ & ADP & VIGwbP\\  
\midrule
adjacent &100 &3 &99.0 &100.0 &100.0 &100.0 \\ 
  &  &5 &100.0 &100.0 &100.0 &100.0 \\ 
  &500 &3 &96.2 &99.2 &98.0 &98.7 \\ 
  &  &5 &98.7 &100.0 &99.8 &99.9 \\ 
  &1000 &3 &90.4 &97.3 &95.4 &93.4 \\ 
  &  &5 &93.4 &99.7 &98.5 &98.0 \\ \hline
random &100 &3 &99.7 &100.0 &100.0 &100.0 \\ 
  &  &5 &100.0 &100.0 &100.0 &100.0 \\ 
  &500 &3 &98.2 &99.9 &99.5 &99.7 \\ 
  &  &5 &98.2 &100.0 &100.0 &100.0 \\ 
  &1000 &3 &91.6 &98.6 &97.2 &96.7 \\ 
  &  &5 &91.0 &99.6 &98.1 &99.1 \\ 
 \bottomrule
\end{tabular}
\label{tab:nedges_NK_fixed_gen}
\end{table} 

\begin{table}[h]
\scriptsize
\centering
\caption{Median of the percentage of the edges of the empirical VIGw over all possible edges for the 0-1 knapsack problem experiment with fixed number of iterations.}
\begin{tabular}{r|rrrr}
  \toprule
$N$ & SRP, $\alpha=2$ & SRP, $\alpha=50$ & ADP & VIGwbP\\  
\midrule
500 &58.3 &61.5 &61.5 &62.9 \\ 
1000 &43.7 &30.6 &28.2 &30.0 \\ 
1500 &36.0 &17.9 &15.3 &17.8 \\ 
2000 &30.5 &11.5 &9.4 &12.4 \\ 
 \bottomrule
\end{tabular}
\label{tab:nedges_KS_fixed_gen}
\end{table}

\begin{table}[h]
\scriptsize
\centering
\caption{Median of the percentage of the edges of the empirical VIGw over all possible edges for the feature selection experiment with fixed number of iterations.}
\begin{tabular}{r|rrrr}
  \toprule
dataset & SRP, $\alpha=2$ & SRP, $\alpha=50$ & ADP & VIGwbP\\  
\midrule
housing &100.0 &100.0 &100.0 &100.0 \\ 
friedman &98.6 &100.0 &100.0 &100.0 \\ 
ionosphere &94.1 &94.1 &94.1 &94.1 \\ 
sonar &99.4 &100.0 &100.0 &99.9 \\ 
cnae9m &39.0 &77.2 &57.9 &58.4 \\ 
libras &95.5 &100.0 &100.0 &99.1 \\ 
covidxr &94.4 &99.8 &99.7 &98.4 \\ 
arrhythmia &33.7 &47.4 &45.7 &43.9 \\   
 \bottomrule
\end{tabular}
\label{tab:nedges_FS_fixed_gen}
\end{table}

\subsection{Results: comparing perturbation strategies}
 \label{sec:res_VIGwbp}
The results for the comparison of the different perturbation strategies can be seen in the last 4 columns of tables~\ref{tab:res_NK_fixed_gen_1}-\ref{tab:res_FS_fixed_gen} (for the experiments with fixed number of iterations) and tables~\ref{tab:res_NK_fixed_time}-\ref{tab:res_FS_fixed_time} (for the experiments with fixed time). 
Tables S10-S15  (for experiments with fixed number of iterations) and S16-S18 (for the experiments with fixed time) of the Supplementary Material show the corrected p-values and the results of the Wilcoxon signed rank test used for the comparison of VIGwbP with the other 3 perturbation strategies. 

\begin{figure}[h]
  \centering
  \includegraphics[width=0.4\linewidth]{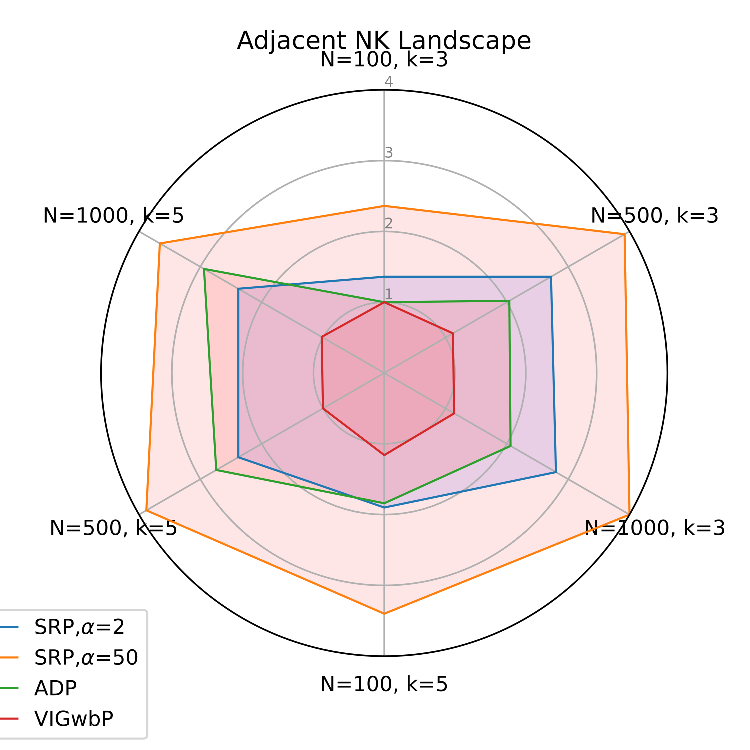}
  \includegraphics[width=0.4\linewidth]{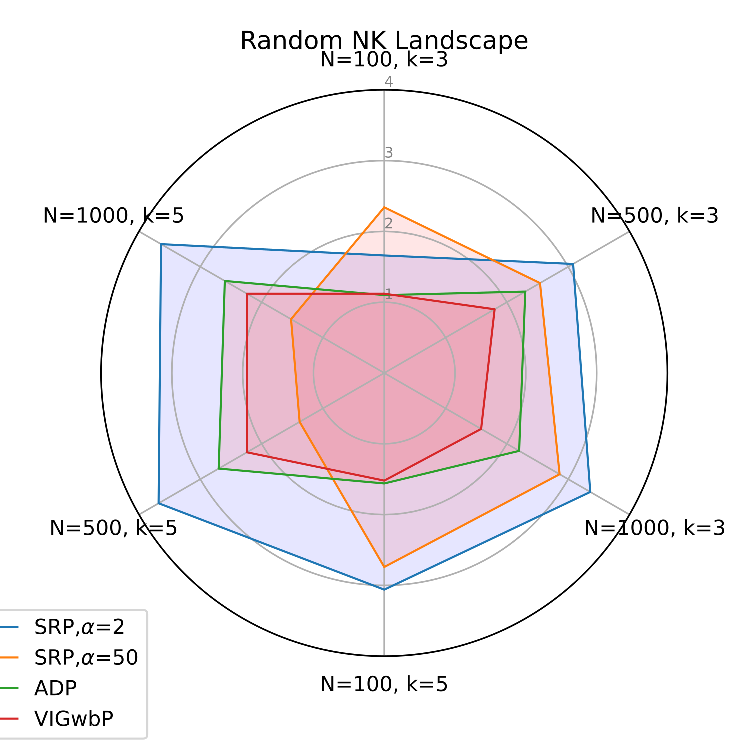} \\
  \includegraphics[width=0.4\linewidth]{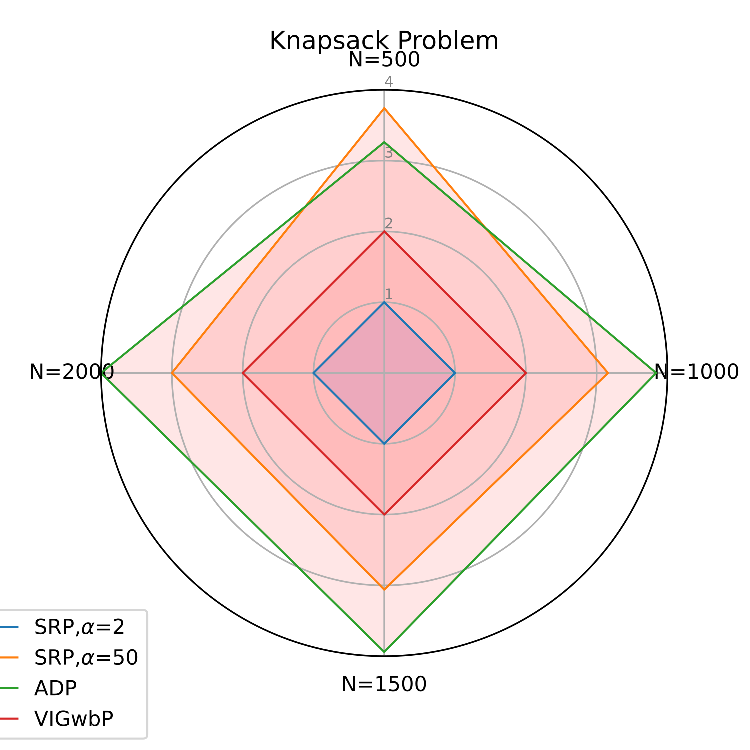}
  \includegraphics[width=0.4\linewidth]{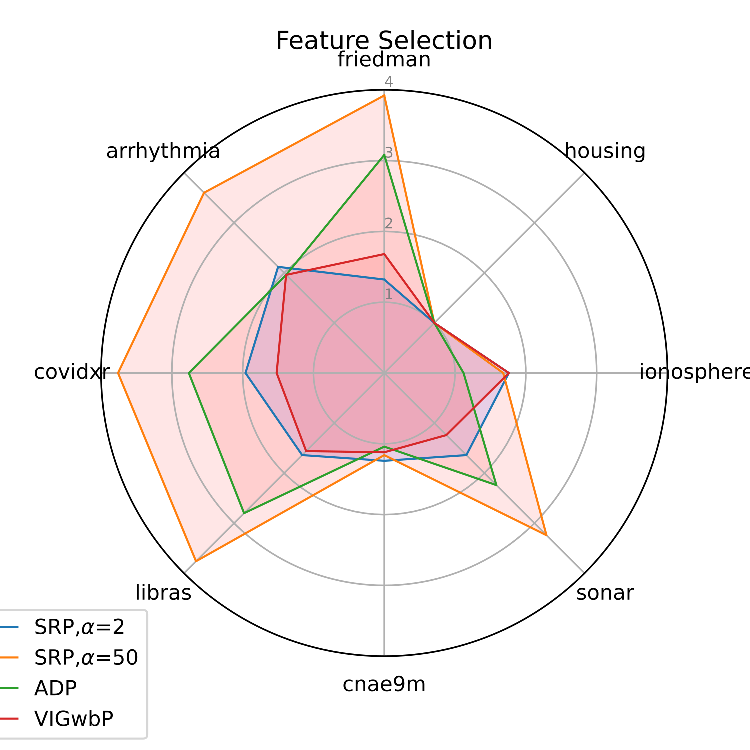}
  \caption{Average ranking for the comparison of $FIT$ for different perturbation strategies in the experiments with fixed number of iterations.} 
  \label{fig:ranking_iterations}
\end{figure}

\begin{figure}[h]
  \centering
  \includegraphics[width=0.4\linewidth]{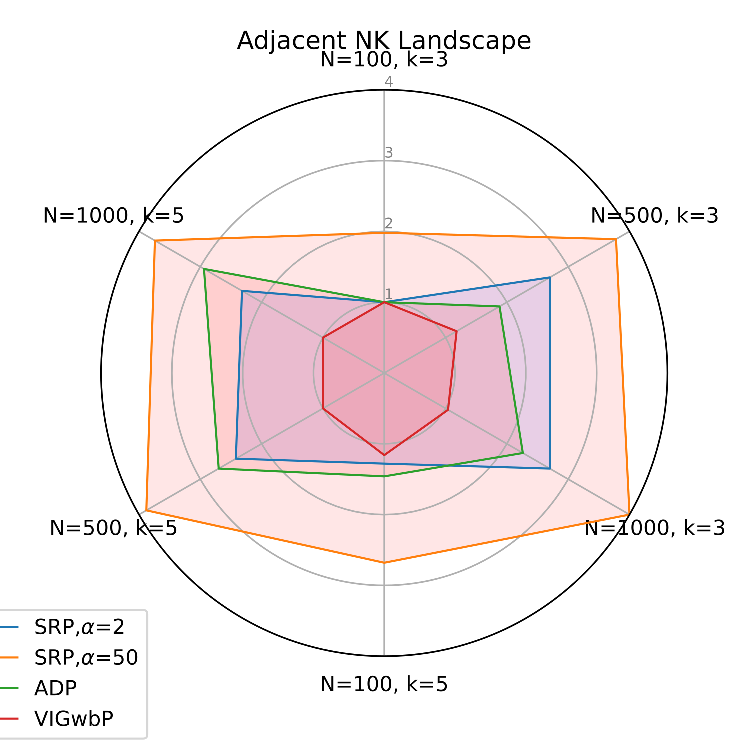}
  \includegraphics[width=0.4\linewidth]{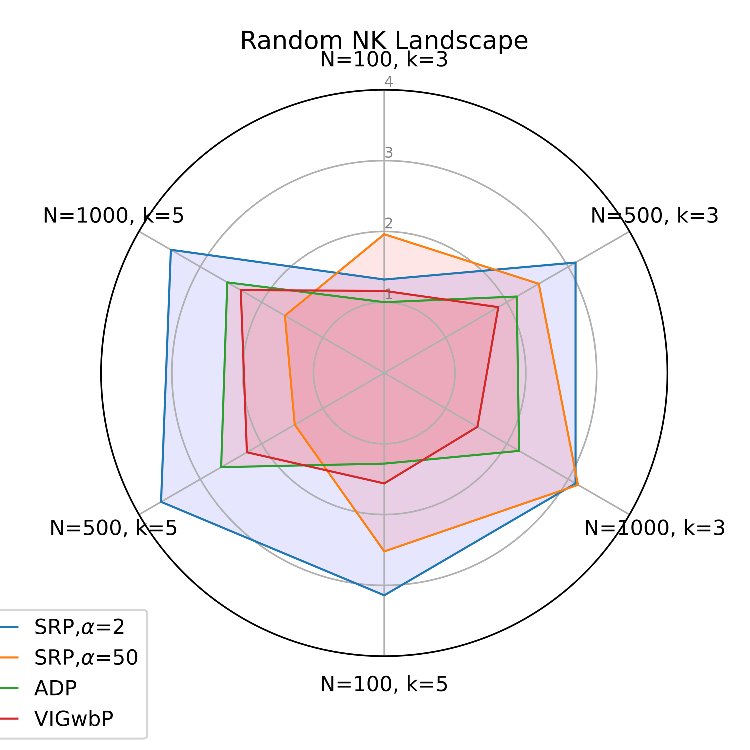} \\
  \includegraphics[width=0.4\linewidth]{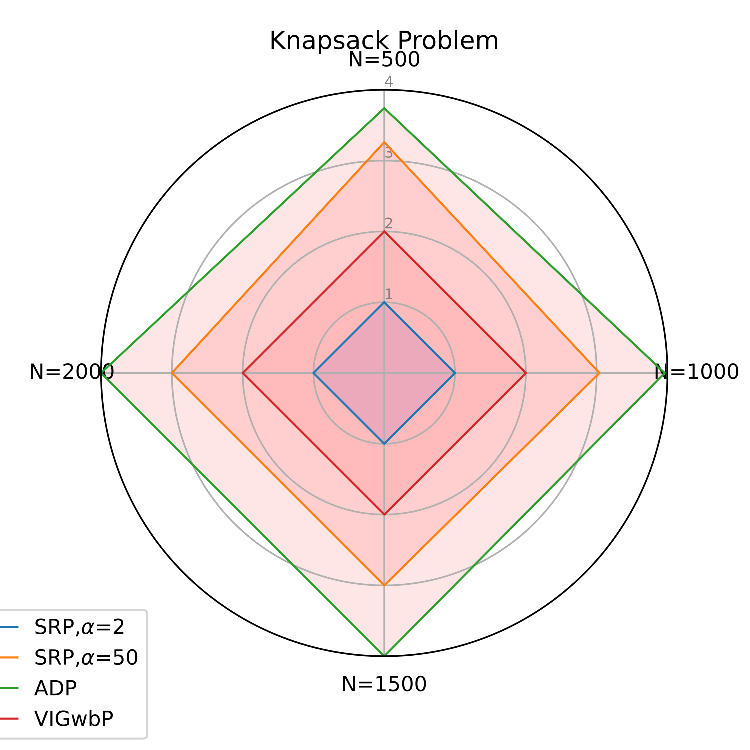}
   \includegraphics[width=0.4\linewidth]{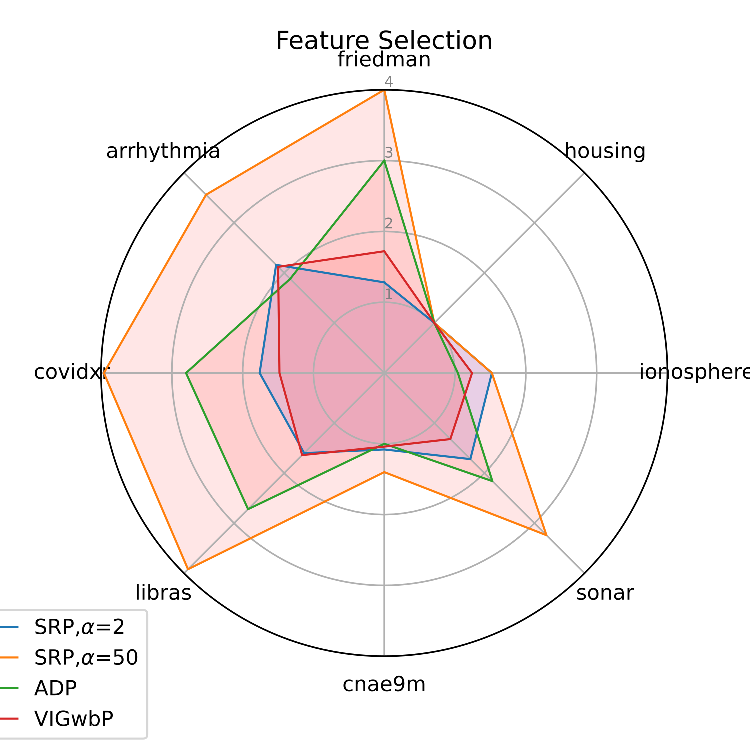}
  \caption{Average ranking for the comparison of $FIT$ for different perturbation strategies in the experiments with fixed time.}
  \label{fig:ranking_time}
\end{figure}

Figures \ref{fig:ranking_iterations} and \ref{fig:ranking_time} show the radar charts of the average rankings~\cite{latorre2021} for the comparison of $FIT$ for different perturbation strategies. 
Each radar chart shows the average rating obtained by the four strategies in experiments with different parameters.
Radar charts make it easy to analyze the impact of parameters on the performance of strategies.
Smaller values for the average rank are better; for example, if the average rank of a strategy is 1.0, it means that the strategy resulted in the best results in all runs of the experiment. 
VIGwbP presented the best average ranking for the adjacent NK landscapes model. 
SRP with $\alpha=50$ presented the worse results. 
For the random NK landscapes model, SRP with $\alpha=50$ presented the best average ranking for large dimension ($N=500$ and $N=1000$) and $k=5$. 
In the other four cases, VIGwbP presented the best ranking. 
In the knapsack and feature selection problems, the  perturbation strategies with best average ranking were SRP with $\alpha=2$ and VIGwbP. 

For NK landscapes, smaller $HDP$ generally implies a smaller Hamming distance between consecutive local optima ($HDLO$) and a smaller number of iterations for the local search functions ($NILS$). 
As a consequence, the runtime ($TIME$) is smaller when the number of iterations is fixed. 
Small jumps can keep the perturbed solution in the same local optimum basin of attraction. 
We can observe this by analyzing  the percentage of escapes from local optimum in ILS ($PELO$). 
Perturbation strategies with averaged smaller perturbation strength presents smaller $PELO$. 

However, the performance of ILS is related not only to the perturbation strength but also to which variables are changed. 
While VIGwbP flips decision variables that interact, the other strategies flip random decision variables. 
Local search systematically flips the variables and, eventually, a flipped variable interacts with the variable changed by perturbation. 
However, interactions of less flipped variables will be associated to the subfunctions when $N$ increases and $k$ is small. 
ILS with VIGwbP generally changes a smaller number of subfunctions, and as a consequence results in smaller $FDP$, for each perturbed variable. 
This can be seen in the ratio $FHRP$. 
ILS with VIGwbP generally presented the best value of $FHRP$ in the experiments with NK landscapes. 
In other words, VIGwbP generally changed less terms of Eq.~\eqref{eq:walsh-decomposition} associated to nonzero Walsh coefficients for a given number of flipped variables. 
Better fitness results were obtained, mainly for the adjacent model, for higher $N$, and for smaller $k$. 
In the adjacent model, the same variables appear in more subfunctions than in the random model. 
This explains the better results in the adjacent model. 
Another advantage of VIGwbP is that the number of flipped variables changes during optimization. 
The number of flipped variables also depends on the epistasis degree and the distribution of the weights in the graph. 
ADP also changes the perturbation strength during optimization. 
However, the results of VIGw are better than the results of ADP, showing that flipping variables that interact impacts positively the performance of ILS. 

In the knapsack problem, the best results are for SRP with $\alpha=2$. 
The smaller number of flipped bits (smaller $HDP$), resulted in smaller $HDLO$. 
Most important, it also resulted in smaller $FDP$. 
This is explained by the penalty function (Eq.~\ref{eq:knap_p}) used in the evaluation of the solutions. 
Flipping more variables in local optima generally results in solutions that exceeds the capacity $C$ of the knapsack. 
However, even flipping in general more bits (higher $HDP$) than SRP with $\alpha=50$, VIGwbP presented better performance and the second best average ranking. 
This again is a direct result of changing variables that strongly interact in VIGwbP.

In the feature selection problem, the best algorithms were again SRP with $\alpha=2$ and VIGwbP. 
VIGwbP presented better average ranking in more datasets, but no statistical difference was observed in the comparison of SRP with $\alpha=2$ and VIGwbP, except for 1 out of 16 times. 
In the experiments, VIGwbP generally flipped less variables (smaller $HDP$) than SRP with $\alpha=50$ and ADP, what impacted positively $HDLO$ and $FDP$. 
HDP is higher for VIGwbP when compared to SRP with $\alpha=2$, but this did not significantly impact $FIT$.

\section{Conclusions}
\label{sec:con}
\citet{tinos2022} proposed local search with linkage
learning (LSwLL) that builds an empirical variable interaction graph (VIG). 
Here, based on LSwLL, we propose LSwLL2, that builds an empirical weighted variable interaction graph (VIGw). 
The VIGw is a weighted undirected graph where the weights represent the strength of the interaction between variables.
Results of experiments with NK landscapes show that LSwLL2 builds empirical VIGws with $90\%$ or more of the edges of the VIG. 
No false linkage is returned. 
It also generally did not significantly affect the performance of the algorithms in the experiments. 
The information of the structure of the problem is obtained as a side effect of local search. 
In addition, the strategy learns information about the strength of the interaction. 
As far as we know, this is new in the literature.

The VIGw produced by LSwLL2 allows us to obtain new insights about the optimization problem and optimizers. 
When compared to the VIG produced by LSwLL, the VIGw produced by LSwLL2 is more informative. 
In addition, it can be useful in problems where the VIG produced by LSwLL is not helpful, e.g., when the VIG is dense. 
Experiments with NK landscapes show that the strongest interactions, detected in the VIGw, occur for variables that impact subfunctions with largest contribution to the evaluation function $f(\mathbf{x})$. 
In experiments with the 0-1 knapsack problem, strongest interactions occur between variables associated to heavy objects because changing these variables together impact more the penalty applied to the evaluation function. 
Finally, experiments with feature selection showed that the VIGw can be used for the visualization of interactions between pairs of variables (features) in machine learning. 
Finding variable interaction is a relevant problem in machine learning. 
Variable interaction information is produced during feature selection in ILS with LSwLL2, without needing additional solution evaluations. 

New transformation operators and optimization strategies that explore the weights of the VIGw produced by LSwLL2 can be designed. 
We illustrate this ability by proposing a perturbation strategy based on the VIGw for ILS.  
In VIGwbP, decision variables with strongest interactions are perturbed together. 
This does not necessarily result in a smaller number of iterations for local search. 
However, it can potentially result in better performance of ILS. 
In the experiments with NK landscapes, the ratio between the number of subfunctions changed by perturbation and the Hamming distance between solutions before and after perturbation was generally smaller for VIGwbP. In other words, the similarity regarding the number of terms associated to nonzero Walsh coefficients (Eq.~\ref{eq:walsh-decomposition}) was higher for the same number of flipped variables. 
Another advantage of VIGwbP is that the number of flipped variables is not fixed and changes with the epistasis degree. VIGwbP is parameterless. 
Good results were also obtained for the 0-1 knapsack problem and feature selection problem. 
However, another perturbation strategy presented better results for 0-1 knapsack problem; VIGwbP presented the second best average ranking. 
This is explained by penalties applied to the evaluation of the solutions. 
It should be interesting to investigate if similar results can be found for different formulations of the knapsack problem, e.g., the quadratic binary unconstrained optimization formulation~\cite{quintero2021}.

In the future, VIGwbP can be used in conjunction with tabu search and other adaptive perturbation strategies (see Section~\ref{sec:VIGwbP}). Investigating variations of VIGwbP for different applications and domains is also attractive. 
In particular, it should be attractive for evolutionary dynamic optimization~\cite{nguyen2012}, where perturbation can assume an important role. 
It should also be attractive to other combinatorial optimization problems, e.g., the MAX-kSAT problem. 
MAX-kSAT has a problem structure similar to NK landscapes; both problems can be seen as Mk landscapes~\cite{whitley2016}. 
In this way, LSwLL2 should be able to discover relevant problem structures for MAX-kSAT. 
Exploring information of the VIGw has potential to result in efficient transformation operators for MAX-kSAT. 
In the past, gray-box optimization operators developed for NK landscapes were proven to be also efficient for MAX-kSAT~\cite{chen2018}. 
The introduction of LSwLL2 into existing evolutionary algorithms has also a significant potential. 
LSwLL2 can be used in black-box optimization problems with strategies developed for efficient gray-box optimization. 
An example is the use of LSwLL2 to build empirical VIGws for partition crossover~\cite{tinos2015,chicano2021} and in \emph{deterministic recombination and iterated local search} (DRILS)~\cite{chicano2017}.

\begin{acks}
This work was partially supported in Brazil by S\~ao Paulo Research Foundation - FAPESP (under grant \#2021/09720-2), National Council for Scientific and Technological Development - CNPq (under grant \#306689/2021-9) and Center for Artificial Intelligence - C4AI (supported by FAPESP, under grant \#2019/07665-4, and IBM Corporation), in Poland by the Polish National Science Centre - NCN (under grant \#2022/45/B/ST6/04150), in Spain by the PID 2020-116727RB-I00 (HUmove) funded by MCIN/AEI/10.13039/501100011033 and TAILOR ICT-48 Network (No 952215) funded by EU Horizon 2020 research and innovation programme.
\end{acks}

\bibliographystyle{ACM-Reference-Format}
\bibliography{LSwLL_2022_f}


\begin{thebibliography}{44}


\ifx \showCODEN    \undefined \def \showCODEN     #1{\unskip}     \fi
\ifx \showDOI      \undefined \def \showDOI       #1{#1}\fi
\ifx \showISBNx    \undefined \def \showISBNx     #1{\unskip}     \fi
\ifx \showISBNxiii \undefined \def \showISBNxiii  #1{\unskip}     \fi
\ifx \showISSN     \undefined \def \showISSN      #1{\unskip}     \fi
\ifx \showLCCN     \undefined \def \showLCCN      #1{\unskip}     \fi
\ifx \shownote     \undefined \def \shownote      #1{#1}          \fi
\ifx \showarticletitle \undefined \def \showarticletitle #1{#1}   \fi
\ifx \showURL      \undefined \def \showURL       {\relax}        \fi
\providecommand\bibfield[2]{#2}
\providecommand\bibinfo[2]{#2}
\providecommand\natexlab[1]{#1}
\providecommand\showeprint[2][]{arXiv:#2}

\bibitem[Andonov et~al\mbox{.}(2000)]%
        {andonov2000}
\bibfield{author}{\bibinfo{person}{R. Andonov}, \bibinfo{person}{V. Poirriez},
  {and} \bibinfo{person}{S. Rajopadhye}.} \bibinfo{year}{2000}\natexlab{}.
\newblock \showarticletitle{Unbounded knapsack problem: Dynamic programming
  revisited}.
\newblock \bibinfo{journal}{\emph{European Journal of Operational Research}}
  \bibinfo{volume}{123}, \bibinfo{number}{2} (\bibinfo{year}{2000}),
  \bibinfo{pages}{394--407}.
\newblock


\bibitem[Battiti and Protasi(1997)]%
        {battiti1997}
\bibfield{author}{\bibinfo{person}{R. Battiti} {and} \bibinfo{person}{M.
  Protasi}.} \bibinfo{year}{1997}\natexlab{}.
\newblock \showarticletitle{Reactive search, a history-based heuristic for
  MAX-SAT}.
\newblock \bibinfo{journal}{\emph{ACM Journal of Experimental Algorithmics}}
  \bibinfo{volume}{2}, \bibinfo{number}{10.1145} (\bibinfo{year}{1997}),
  \bibinfo{pages}{264216--264220}.
\newblock


\bibitem[Bosman et~al\mbox{.}(2016)]%
        {bosman2016}
\bibfield{author}{\bibinfo{person}{P.~A.~N. Bosman}, \bibinfo{person}{N.~H.
  Luong}, {and} \bibinfo{person}{D. Thierens}.}
  \bibinfo{year}{2016}\natexlab{}.
\newblock In \bibinfo{booktitle}{\emph{Proceedings of the Genetic and
  Evolutionary Computation Conference}}. \bibinfo{pages}{637--644}.
\newblock


\bibitem[Brand{\~a}o(2020)]%
        {brandao2020}
\bibfield{author}{\bibinfo{person}{J. Brand{\~a}o}.}
  \bibinfo{year}{2020}\natexlab{}.
\newblock \showarticletitle{A memory-based iterated local search algorithm for
  the multi-depot open vehicle routing problem}.
\newblock \bibinfo{journal}{\emph{European Journal of Operational Research}}
  \bibinfo{volume}{284}, \bibinfo{number}{2} (\bibinfo{year}{2020}),
  \bibinfo{pages}{559--571}.
\newblock


\bibitem[Chen et~al\mbox{.}(2018)]%
        {chen2018}
\bibfield{author}{\bibinfo{person}{W. Chen}, \bibinfo{person}{D. Whitley},
  \bibinfo{person}{R. Tin{\'o}s}, {and} \bibinfo{person}{F. Chicano}.}
  \bibinfo{year}{2018}\natexlab{}.
\newblock \showarticletitle{Tunneling between plateaus: improving on a
  state-of-the-art MAXSAT solver using partition crossover}. In
  \bibinfo{booktitle}{\emph{Proceedings of the Genetic and Evolutionary
  Computation Conference}}. \bibinfo{pages}{921--928}.
\newblock


\bibitem[Chicano et~al\mbox{.}(2021)]%
        {chicano2021}
\bibfield{author}{\bibinfo{person}{F. Chicano}, \bibinfo{person}{G. Ochoa},
  \bibinfo{person}{D. Whitley}, {and} \bibinfo{person}{R. Tin{\'o}s}.}
  \bibinfo{year}{2021}\natexlab{}.
\newblock \showarticletitle{Dynastic Potential Crossover Operator}.
\newblock \bibinfo{journal}{\emph{Evolutionary Computation}}
  \bibinfo{volume}{online first} (\bibinfo{year}{2021}),
  \bibinfo{pages}{1--35}.
\newblock


\bibitem[Chicano et~al\mbox{.}(2017)]%
        {chicano2017}
\bibfield{author}{\bibinfo{person}{F. Chicano}, \bibinfo{person}{D. Whitley},
  \bibinfo{person}{G. Ochoa}, {and} \bibinfo{person}{R. Tin{\'o}s}.}
  \bibinfo{year}{2017}\natexlab{}.
\newblock \showarticletitle{Optimizing one million variable NK landscapes by
  hybridizing deterministic recombination and local search}. In
  \bibinfo{booktitle}{\emph{Proceedings of the Genetic and Evolutionary
  Computation Conference}}. \bibinfo{pages}{753--760}.
\newblock


\bibitem[Chicano et~al\mbox{.}(2014)]%
        {chicano2014}
\bibfield{author}{\bibinfo{person}{F. Chicano}, \bibinfo{person}{D. Whitley},
  {and} \bibinfo{person}{A.~M. Sutton}.} \bibinfo{year}{2014}\natexlab{}.
\newblock \showarticletitle{Efficient identification of improving moves in a
  ball for pseudo-boolean problems}. In \bibinfo{booktitle}{\emph{Proceedings
  of the Genetic and Evolutionary Computation Conference}}.
  \bibinfo{pages}{437--444}.
\newblock


\bibitem[Coffin and Clack(2006)]%
        {coffin2006}
\bibfield{author}{\bibinfo{person}{D.~J. Coffin} {and} \bibinfo{person}{C.~D.
  Clack}.} \bibinfo{year}{2006}\natexlab{}.
\newblock \showarticletitle{gLINC: Identifying Composability Using Group
  Perturbation}. In \bibinfo{booktitle}{\emph{Proceedings of the Genetic and
  Evolutionary Computation Conference}}. \bibinfo{pages}{1133--1140}.
\newblock


\bibitem[Dowsland and Thompson(2012)]%
        {dowsland2012}
\bibfield{author}{\bibinfo{person}{K.~A. Dowsland} {and} \bibinfo{person}{J.
  Thompson}.} \bibinfo{year}{2012}\natexlab{}.
\newblock \showarticletitle{Simulated Annealing}.
\newblock In \bibinfo{booktitle}{\emph{Handbook of Natural Computing}}.
  \bibinfo{publisher}{Springer-Verlag}, \bibinfo{pages}{1623--1655}.
\newblock


\bibitem[Dua and Graff(2017)]%
        {dua2019}
\bibfield{author}{\bibinfo{person}{D. Dua} {and} \bibinfo{person}{C. Graff}.}
  \bibinfo{year}{2017}\natexlab{}.
\newblock \bibinfo{title}{{UCI} Machine Learning Repository}.
\newblock
\newblock
\urldef\tempurl%
\url{http://archive.ics.uci.edu/ml}
\showURL{%
\tempurl}


\bibitem[Friedman and Popescu(2008)]%
        {friedman2008}
\bibfield{author}{\bibinfo{person}{J.~H. Friedman} {and} \bibinfo{person}{B.~E.
  Popescu}.} \bibinfo{year}{2008}\natexlab{}.
\newblock \showarticletitle{Predictive learning via rule ensembles}.
\newblock \bibinfo{journal}{\emph{The Annals of Applied Statistics}}
  (\bibinfo{year}{2008}), \bibinfo{pages}{916--954}.
\newblock


\bibitem[Goldman and Punch(2014)]%
        {goldman2014}
\bibfield{author}{\bibinfo{person}{B.~W. Goldman} {and} \bibinfo{person}{W.~F.
  Punch}.} \bibinfo{year}{2014}\natexlab{}.
\newblock \showarticletitle{Parameter-less Population Pyramid}. In
  \bibinfo{booktitle}{\emph{Proceedings of the Genetic and Evolutionary
  Computation Conference}}. \bibinfo{pages}{785--792}.
\newblock


\bibitem[Han and Kim(2000)]%
        {han2000}
\bibfield{author}{\bibinfo{person}{K.-H. Han} {and} \bibinfo{person}{J.-H.
  Kim}.} \bibinfo{year}{2000}\natexlab{}.
\newblock \showarticletitle{Genetic quantum algorithm and its application to
  combinatorial optimization problem}. In \bibinfo{booktitle}{\emph{Proceedings
  of the IEEE Congress on Evolutionary Computation}}, Vol.~\bibinfo{volume}{2}.
  \bibinfo{pages}{1354--1360}.
\newblock


\bibitem[Hansen and Mladenovi{\'c}(2003)]%
        {hansen2003}
\bibfield{author}{\bibinfo{person}{P. Hansen} {and} \bibinfo{person}{N.
  Mladenovi{\'c}}.} \bibinfo{year}{2003}\natexlab{}.
\newblock \showarticletitle{Variable neighborhood search}.
\newblock In \bibinfo{booktitle}{\emph{Handbook of Metaheuristics}}.
  \bibinfo{publisher}{Springer}, \bibinfo{pages}{145--184}.
\newblock


\bibitem[Heckendorn(2002)]%
        {heckendorn2002}
\bibfield{author}{\bibinfo{person}{R.~B. Heckendorn}.}
  \bibinfo{year}{2002}\natexlab{}.
\newblock \showarticletitle{Embedded Landscapes}.
\newblock \bibinfo{journal}{\emph{Evolutionary Computation}}
  \bibinfo{volume}{10}, \bibinfo{number}{4} (\bibinfo{year}{2002}),
  \bibinfo{pages}{345--369}.
\newblock


\bibitem[Heckendorn and Wright(2004)]%
        {heckendorn2004}
\bibfield{author}{\bibinfo{person}{R.~B. Heckendorn} {and}
  \bibinfo{person}{A.~H. Wright}.} \bibinfo{year}{2004}\natexlab{}.
\newblock \showarticletitle{Efficient linkage discovery by limited probing}.
\newblock \bibinfo{journal}{\emph{Evolutionary Computation}}
  \bibinfo{volume}{12}, \bibinfo{number}{4} (\bibinfo{year}{2004}),
  \bibinfo{pages}{517--545}.
\newblock


\bibitem[Hsu and Yu(2015)]%
        {hsu2015}
\bibfield{author}{\bibinfo{person}{S.-H. Hsu} {and} \bibinfo{person}{T.-L.
  Yu}.} \bibinfo{year}{2015}\natexlab{}.
\newblock \showarticletitle{Optimization by Pairwise Linkage Detection,
  Incremental Linkage Set, and Restricted / Back Mixing: {DSMGA}-{II}}. In
  \bibinfo{booktitle}{\emph{Proceedings of the Genetic and Evolutionary
  Computation Conference}}. \bibinfo{pages}{519--526}.
\newblock


\bibitem[Inglis et~al\mbox{.}(2022)]%
        {inglis2022}
\bibfield{author}{\bibinfo{person}{A. Inglis}, \bibinfo{person}{A. Parnell},
  {and} \bibinfo{person}{C.~B. Hurley}.} \bibinfo{year}{2022}\natexlab{}.
\newblock \showarticletitle{Visualizing Variable Importance and Variable
  Interaction Effects in Machine Learning Models}.
\newblock \bibinfo{journal}{\emph{Journal of Computational and Graphical
  Statistics}} (\bibinfo{year}{2022}), \bibinfo{pages}{1--13}.
\newblock


\bibitem[LaTorre et~al\mbox{.}(2021)]%
        {latorre2021}
\bibfield{author}{\bibinfo{person}{A. LaTorre}, \bibinfo{person}{D. Molina},
  \bibinfo{person}{E. Osaba}, \bibinfo{person}{J. Poyatos}, \bibinfo{person}{J.
  Del~Ser}, {and} \bibinfo{person}{F. Herrera}.}
  \bibinfo{year}{2021}\natexlab{}.
\newblock \showarticletitle{A prescription of methodological guidelines for
  comparing bio-inspired optimization algorithms}.
\newblock \bibinfo{journal}{\emph{Swarm and Evolutionary Computation}}
  \bibinfo{volume}{67} (\bibinfo{year}{2021}), \bibinfo{pages}{100973}.
\newblock


\bibitem[Li et~al\mbox{.}(2013)]%
        {li2013}
\bibfield{author}{\bibinfo{person}{X. Li}, \bibinfo{person}{K. Tang},
  \bibinfo{person}{M.~N. Omidvar}, \bibinfo{person}{Z. Yang}, {and}
  \bibinfo{person}{K. Qin}.} \bibinfo{year}{2013}\natexlab{}.
\newblock \bibinfo{booktitle}{\emph{Benchmark functions for the CEC 2013
  special session and competition on large-scale global optimization}}.
\newblock \bibinfo{type}{{T}echnical {R}eport}. \bibinfo{institution}{RMIT
  University, Australia}.
\newblock


\bibitem[Louren{\c{c}}o et~al\mbox{.}(2019)]%
        {lourenco2019}
\bibfield{author}{\bibinfo{person}{H.~R. Louren{\c{c}}o},
  \bibinfo{person}{O.~C. Martin}, {and} \bibinfo{person}{T. St{\"u}tzle}.}
  \bibinfo{year}{2019}\natexlab{}.
\newblock \showarticletitle{Iterated local search: Framework and applications}.
\newblock In \bibinfo{booktitle}{\emph{Handbook of Metaheuristics}}.
  \bibinfo{publisher}{Springer}, \bibinfo{pages}{129--168}.
\newblock


\bibitem[L{\"u} and Hao(2009)]%
        {lu2009}
\bibfield{author}{\bibinfo{person}{Z. L{\"u}} {and} \bibinfo{person}{J.-K.
  Hao}.} \bibinfo{year}{2009}\natexlab{}.
\newblock \showarticletitle{A critical element-guided perturbation strategy for
  iterated local search}. In \bibinfo{booktitle}{\emph{European Conference on
  Evolutionary Computation in Combinatorial Optimization}}. Springer,
  \bibinfo{pages}{1--12}.
\newblock


\bibitem[Nguyen et~al\mbox{.}(2012)]%
        {nguyen2012}
\bibfield{author}{\bibinfo{person}{T.~T. Nguyen}, \bibinfo{person}{S. Yang},
  {and} \bibinfo{person}{J. Branke}.} \bibinfo{year}{2012}\natexlab{}.
\newblock \showarticletitle{Evolutionary dynamic optimization: A survey of the
  state of the art}.
\newblock \bibinfo{journal}{\emph{Swarm and Evolutionary Computation}}
  \bibinfo{volume}{6} (\bibinfo{year}{2012}), \bibinfo{pages}{1--24}.
\newblock


\bibitem[Ochoa et~al\mbox{.}(2008)]%
        {ochoa2008}
\bibfield{author}{\bibinfo{person}{G. Ochoa}, \bibinfo{person}{M. Tomassini},
  \bibinfo{person}{S. V{\'e}rel}, {and} \bibinfo{person}{C. Darabos}.}
  \bibinfo{year}{2008}\natexlab{}.
\newblock \showarticletitle{A study of NK landscapes' basins and local optima
  networks}. In \bibinfo{booktitle}{\emph{Proceedings of the Genetic and
  Evolutionary Computation Conference}}. \bibinfo{pages}{555--562}.
\newblock


\bibitem[Papadimitriou and Steiglitz(1998)]%
        {papadimitriou1998}
\bibfield{author}{\bibinfo{person}{C.~H. Papadimitriou} {and}
  \bibinfo{person}{K. Steiglitz}.} \bibinfo{year}{1998}\natexlab{}.
\newblock \bibinfo{booktitle}{\emph{Combinatorial optimization: algorithms and
  complexity}}.
\newblock \bibinfo{publisher}{Dover Publications}.
\newblock


\bibitem[Przewozniczek et~al\mbox{.}(2020)]%
        {przewozniczek2020b}
\bibfield{author}{\bibinfo{person}{M.~W. Przewozniczek}, \bibinfo{person}{B.
  Frej}, {and} \bibinfo{person}{M.~M. Komarnicki}.}
  \bibinfo{year}{2020}\natexlab{}.
\newblock \showarticletitle{On Measuring and Improving the Quality of Linkage
  Learning in Modern Evolutionary Algorithms Applied to Solve Partially
  Additively Separable Problems}. In \bibinfo{booktitle}{\emph{Proceedings of
  the Genetic and Evolutionary Computation Conference}}.
  \bibinfo{pages}{742–750}.
\newblock


\bibitem[Przewozniczek and Komarnicki(2020)]%
        {przewozniczek2020}
\bibfield{author}{\bibinfo{person}{M.~W. Przewozniczek} {and}
  \bibinfo{person}{M.~M. Komarnicki}.} \bibinfo{year}{2020}\natexlab{}.
\newblock \showarticletitle{Empirical Linkage Learning}.
\newblock \bibinfo{journal}{\emph{IEEE Transactions on Evolutionary
  Computation}} \bibinfo{volume}{24}, \bibinfo{number}{6}
  (\bibinfo{year}{2020}), \bibinfo{pages}{1097--1111}.
\newblock


\bibitem[Przewozniczek et~al\mbox{.}(2021)]%
        {przewozniczek2021}
\bibfield{author}{\bibinfo{person}{M.~W. Przewozniczek}, \bibinfo{person}{M.~M.
  Komarnicki}, {and} \bibinfo{person}{B. Frej}.}
  \bibinfo{year}{2021}\natexlab{}.
\newblock \showarticletitle{Direct linkage discovery with empirical linkage
  learning}. In \bibinfo{booktitle}{\emph{Proceedings of the Genetic and
  Evolutionary Computation Conference}}. \bibinfo{pages}{609--617}.
\newblock


\bibitem[Przewozniczek et~al\mbox{.}(2022)]%
        {przewozniczek2022}
\bibfield{author}{\bibinfo{person}{M.~W. Przewozniczek}, \bibinfo{person}{R.
  Tin\'{o}s}, \bibinfo{person}{B. Frej}, {and} \bibinfo{person}{M.~M.
  Komarnicki}.} \bibinfo{year}{2022}\natexlab{}.
\newblock \showarticletitle{On Turning Black - into Dark Gray-Optimization with
  the Direct Empirical Linkage Discovery and Partition Crossover}. In
  \bibinfo{booktitle}{\emph{Proceedings of the Genetic and Evolutionary
  Computation Conference}}. \bibinfo{pages}{269–277}.
\newblock


\bibitem[Quintero and Zuluaga(2021)]%
        {quintero2021}
\bibfield{author}{\bibinfo{person}{R.~A. Quintero} {and} \bibinfo{person}{L.~F
  Zuluaga}.} \bibinfo{year}{2021}\natexlab{}.
\newblock \bibinfo{booktitle}{\emph{Characterizing and Benchmarking QUBO
  Reformulations of the Knapsack Problem}}.
\newblock \bibinfo{type}{{T}echnical {R}eport}.
  \bibinfo{institution}{Department of Industrial and Systems Engineering,
  Lehigh}.
\newblock


\bibitem[Thierens and Bosman(2012)]%
        {thierens2012}
\bibfield{author}{\bibinfo{person}{D. Thierens} {and} \bibinfo{person}{P.~A.~N.
  Bosman}.} \bibinfo{year}{2012}\natexlab{}.
\newblock \showarticletitle{Predetermined versus Learned Linkage Models}. In
  \bibinfo{booktitle}{\emph{Proceedings of the Genetic and Evolutionary
  Computation Conference}}. \bibinfo{pages}{289–296}.
\newblock


\bibitem[Thierens and Bosman(2013)]%
        {thierens2013}
\bibfield{author}{\bibinfo{person}{D. Thierens} {and} \bibinfo{person}{P.~A.~N.
  Bosman}.} \bibinfo{year}{2013}\natexlab{}.
\newblock \showarticletitle{Hierarchical Problem Solving with the Linkage Tree
  Genetic Algorithm}. In \bibinfo{booktitle}{\emph{Proceedings of the Genetic
  and Evolutionary Computation Conference}}. \bibinfo{pages}{877--884}.
\newblock


\bibitem[Tin{\'o}s(2020)]%
        {tinos2020}
\bibfield{author}{\bibinfo{person}{R. Tin{\'o}s}.}
  \bibinfo{year}{2020}\natexlab{}.
\newblock \showarticletitle{Artificial neural network based crossover for
  evolutionary algorithms}.
\newblock \bibinfo{journal}{\emph{Applied Soft Computing}}
  \bibinfo{volume}{95} (\bibinfo{year}{2020}), \bibinfo{pages}{106512}.
\newblock


\bibitem[Tin{\'o}s et~al\mbox{.}(2022)]%
        {tinos2022}
\bibfield{author}{\bibinfo{person}{R. Tin{\'o}s}, \bibinfo{person}{M.~W.
  Przewozniczek}, {and} \bibinfo{person}{D. Whitley}.}
  \bibinfo{year}{2022}\natexlab{}.
\newblock \showarticletitle{Iterated local search with perturbation based on
  variables interaction for pseudo-boolean optimization}. In
  \bibinfo{booktitle}{\emph{Proceedings of the Genetic and Evolutionary
  Computation Conference}}. \bibinfo{pages}{296--304}.
\newblock


\bibitem[Tin{\'o}s et~al\mbox{.}(2015)]%
        {tinos2015}
\bibfield{author}{\bibinfo{person}{R. Tin{\'o}s}, \bibinfo{person}{D. Whitley},
  {and} \bibinfo{person}{F. Chicano}.} \bibinfo{year}{2015}\natexlab{}.
\newblock \showarticletitle{Partition crossover for pseudo-boolean
  optimization}. In \bibinfo{booktitle}{\emph{Proceedings of the 2015 ACM
  Conference on Foundations of Genetic Algorithms XIII}}.
  \bibinfo{pages}{137--149}.
\newblock


\bibitem[Tin{\'o}s et~al\mbox{.}(2021)]%
        {tinos2021}
\bibfield{author}{\bibinfo{person}{R. Tin{\'o}s}, \bibinfo{person}{D. Whitley},
  \bibinfo{person}{F. Chicano}, {and} \bibinfo{person}{G. Ochoa}.}
  \bibinfo{year}{2021}\natexlab{}.
\newblock \showarticletitle{Partition Crossover for Continuous Optimization:
  ePX}. In \bibinfo{booktitle}{\emph{Proceedings of the Genetic and
  Evolutionary Computation Conference}}. \bibinfo{pages}{627--635}.
\newblock


\bibitem[Van~Griethuysen et~al\mbox{.}(2017)]%
        {van2017}
\bibfield{author}{\bibinfo{person}{J.~J.~M. Van~Griethuysen},
  \bibinfo{person}{A. Fedorov}, \bibinfo{person}{C. Parmar},
  \bibinfo{person}{A. Hosny}, \bibinfo{person}{N. Aucoin}, \bibinfo{person}{V.
  Narayan}, \bibinfo{person}{R.~G.~H. Beets-Tan}, \bibinfo{person}{J.-C.
  Fillion-Robin}, \bibinfo{person}{S. Pieper}, {and} \bibinfo{person}{H.~J.
  W.~L. Aerts}.} \bibinfo{year}{2017}\natexlab{}.
\newblock \showarticletitle{Computational radiomics system to decode the
  radiographic phenotype}.
\newblock \bibinfo{journal}{\emph{Cancer Research}} \bibinfo{volume}{77},
  \bibinfo{number}{21} (\bibinfo{year}{2017}), \bibinfo{pages}{e104--e107}.
\newblock


\bibitem[Wang et~al\mbox{.}(2020)]%
        {wang2020}
\bibfield{author}{\bibinfo{person}{L. Wang}, \bibinfo{person}{Z.~Q. Lin}, {and}
  \bibinfo{person}{A. Wong}.} \bibinfo{year}{2020}\natexlab{}.
\newblock \showarticletitle{Covid-net: A tailored deep convolutional neural
  network design for detection of covid-19 cases from chest x-ray images}.
\newblock \bibinfo{journal}{\emph{Scientific Reports}} \bibinfo{volume}{10},
  \bibinfo{number}{1} (\bibinfo{year}{2020}), \bibinfo{pages}{1--12}.
\newblock


\bibitem[Whitley(2019)]%
        {whitley2019}
\bibfield{author}{\bibinfo{person}{D. Whitley}.}
  \bibinfo{year}{2019}\natexlab{}.
\newblock \showarticletitle{Next generation genetic algorithms: a user’s
  guide and tutorial}.
\newblock In \bibinfo{booktitle}{\emph{Handbook of Metaheuristics}}.
  \bibinfo{publisher}{Springer}, \bibinfo{pages}{245--274}.
\newblock


\bibitem[Whitley et~al\mbox{.}(2016)]%
        {whitley2016}
\bibfield{author}{\bibinfo{person}{D. Whitley}, \bibinfo{person}{F. Chicano},
  {and} \bibinfo{person}{B.~W. Goldman}.} \bibinfo{year}{2016}\natexlab{}.
\newblock \showarticletitle{Gray box optimization for Mk landscapes (NK
  landscapes and MAX-kSAT)}.
\newblock \bibinfo{journal}{\emph{Evolutionary Computation}}
  \bibinfo{volume}{24}, \bibinfo{number}{3} (\bibinfo{year}{2016}),
  \bibinfo{pages}{491--519}.
\newblock


\bibitem[Williamson et~al\mbox{.}(1989)]%
        {williamson1989}
\bibfield{author}{\bibinfo{person}{D.~F. Williamson}, \bibinfo{person}{R.~A.
  Parker}, {and} \bibinfo{person}{J.~S. Kendrick}.}
  \bibinfo{year}{1989}\natexlab{}.
\newblock \showarticletitle{The box plot: a simple visual method to interpret
  data}.
\newblock \bibinfo{journal}{\emph{Annals of Internal Medicine}}
  \bibinfo{volume}{110}, \bibinfo{number}{11} (\bibinfo{year}{1989}),
  \bibinfo{pages}{916--921}.
\newblock


\bibitem[Wright et~al\mbox{.}(2000)]%
        {wright2000}
\bibfield{author}{\bibinfo{person}{A.~H. Wright}, \bibinfo{person}{R.~K.
  Thompson}, {and} \bibinfo{person}{J. Zhang}.}
  \bibinfo{year}{2000}\natexlab{}.
\newblock \showarticletitle{The computational complexity of NK fitness
  functions}.
\newblock \bibinfo{journal}{\emph{IEEE Transactions on Evolutionary
  Computation}} \bibinfo{volume}{4}, \bibinfo{number}{4}
  (\bibinfo{year}{2000}), \bibinfo{pages}{373--379}.
\newblock


\bibitem[Xue et~al\mbox{.}(2016)]%
        {xue2016}
\bibfield{author}{\bibinfo{person}{B. Xue}, \bibinfo{person}{M. Zhang},
  \bibinfo{person}{W.~N. Browne}, {and} \bibinfo{person}{X. Yao}.}
  \bibinfo{year}{2016}\natexlab{}.
\newblock \showarticletitle{A survey on evolutionary computation approaches to
  feature selection}.
\newblock \bibinfo{journal}{\emph{IEEE Transactions on Evolutionary
  Computation}} \bibinfo{volume}{20}, \bibinfo{number}{4}
  (\bibinfo{year}{2016}), \bibinfo{pages}{606--626}.
\newblock


\end{thebibliography}

\end{document}